\documentclass[journal]{IEEEtran}
\usepackage{cite}
\usepackage{amsmath,amssymb,amsfonts}
\usepackage{algorithmic}
\usepackage{graphicx}
\usepackage{textcomp}

\usepackage{bm}

\usepackage[T1]{fontenc}
\usepackage{hyperref, times, cite}
\usepackage{url}
\usepackage{enumitem}
\usepackage{mathtools}
\usepackage{bm, bbm}
\usepackage{cool}
\usepackage{amssymb,amsfonts,amsmath,amsthm,amscd,dsfont,mathrsfs}
\usepackage{graphicx,float,psfrag,epsfig,amssymb}
\usepackage{wrapfig}
\usepackage{relsize}
\usepackage{color}
\usepackage{pict2e}
\usepackage{algorithm}
\usepackage{caption}
\usepackage{nameref}
\usepackage{enumerate}
\usepackage{stackrel}

\usepackage{caption}
\usepackage{subcaption}

\usepackage{amsthm}

\begin{document}
\renewcommand{\algorithmicrequire}{\textbf{Input:}}
\renewcommand{\algorithmicensure}{\textbf{Output:}}
\thispagestyle{empty}

\title{Learning Multi-Attribute Differential Graphs with Non-Convex Penalties}
\author{ Jitendra K.\ Tugnait    
\thanks{J.K.\ Tugnait is with the Department of 
Electrical \& Computer Engineering,
200 Broun Hall, Auburn University, Auburn, AL 36849, USA. 
Email: tugnajk@auburn.edu . }

\thanks{This work was supported by the National Science Foundation Grant CCF-2308473.}}

\maketitle

\begin{abstract}
We consider the problem of estimating differences in two multi-attribute Gaussian graphical models (GGMs) which are known to have similar structure, using a penalized D-trace loss function with non-convex penalties. The GGM structure is encoded in its precision (inverse covariance) matrix. Existing methods for multi-attribute differential graph estimation are based on a group lasso penalized loss function. In this paper, we consider a penalized D-trace loss function with non-convex (log-sum and smoothly clipped absolute deviation (SCAD)) penalties. Two proximal gradient descent methods are presented to optimize the objective function. Theoretical analysis establishing sufficient conditions for consistency in support recovery, convexity and estimation in high-dimensional settings is provided. We illustrate our approaches with numerical examples based on synthetic and real data.
\end{abstract}


\section{Introduction} \label{intro}
\IEEEPARstart{G}{raphical} models are used to display and explore conditional independence structure of the random variables in a system \cite{Whittaker1990, Lauritzen1996,  Buhlmann2011}. The conditional statistical dependency structure among $p$ random variables $x_1, x_2, \cdots , x_p$, is represented using an undirected graph ${\cal G} = \left( V, {\cal E} \right)$ with a set of $p$ vertices (nodes) $V = \{1,2, \cdots , p\} =[p]$, and a corresponding set of (undirected) edges ${\cal E} \subseteq [p] \times [p]$. The graph ${\cal G}$ is a conditional independence graph (CIG) where there is no edge between nodes $i$ and $j$ if and only if (iff) $x_i$ and $x_j$ are conditionally independent given the remaining $p$-$2$ variables. Gaussian graphical models (GGMs) are CIGs where ${\bm x}$ is multivariate Gaussian and $\{i,j\} \not\in {\cal E}$ iff $[\bm{\Omega}]_{ij} =0$ (where for zero-mean ${\bm x}$, precision matrix $\bm{\Omega} = (E \{ {\bm x} {\bm x}^\top \} )^{-1}$). There is extensive literature on this topic \cite{Lauritzen1996,  Buhlmann2011}.. Given $n$ samples of ${\bm x}$, in high-dimensional settings ($n$ is of the order of $p$), one estimates $\bm{\Omega}$ under some sparsity constraints, see, e.g., \cite{Buhlmann2011, Lam2009, Ravikumar2011, Wainwright2019}. 

In differential network analysis one is interested in estimating the difference in two inverse covariance matrices \cite{Yuan2017, Tang2020, Wu2020}. Given observations ${\bm x}$ and ${\bm y}$ from two groups of subjects, one is interested in the difference ${\bm \Delta} = {\bm \Omega}_y - {\bm \Omega}_x$, where ${\bm \Omega}_x = (E \{ {\bm x} {\bm x}^\top \} )^{-1}$ and ${\bm \Omega}_y = (E \{ {\bm y} {\bm y}^\top \} )^{-1}$. The associated differential graph is ${\cal G}_\Delta = \left( V, {\cal E}_\Delta \right)$ where $\{i,j\} \in {\cal E}_\Delta$ iff $[{\bm \Delta}]_{ij} \ne 0$. In biostatistics, the differential network/graph describes the changes in conditional dependencies between components under different environmental or genetic conditions \cite{Tang2020, Zhao2014, Belilovsky2016}. Given gene expression data or functional MRI signals, one is interested in the differences in the graphical models of healthy and impaired subjects, or models under different disease states \cite{Danaher2014, Zhao2014, Belilovsky2016}.

In many applications, there may be more than one random variable associated with a node. This class of graphical models has been called multi-attribute graphical models in \cite{Kolar2014, Tugnait21a, Tugnait2024} and vector graphs or networks in \cite{Marjanovic18, Sundaram20}. Consider $p$ jointly Gaussian vectors ${\bm z}_i \in \mathbb{R}^m$, $i \in [p]$. We associate ${\bm z}_i$ with the $i$th node of graph ${\cal G} = \left( V, {\cal E} \right)$,  $V = [p]$, ${\cal E} \subseteq V \times V$. We now have $m$ attributes per node. Now $\{ i,j \} \in {\cal E}$ iff vectors ${\bm z}_i$ and ${\bm z}_j$ are conditionally independent given the remaining $p$-$2$ vectors $\{ {\bm z}_\ell \, , \ell \in V \textbackslash \{i_, j \} \}$. Let ${\bm x} = [ {\bm z}_1^\top \; {\bm z}_2^\top \; \cdots \; {\bm z}_p^\top ]^\top \in \mathbb{R}^{mp}$. 
Let ${\bm \Omega} = (E \{ {\bm x} {\bm x}^\top \} )^{-1}$ and define the $m \times m$ subblock ${\bm \Omega}^{(ij)}$ of ${\bm \Omega}$ as 
\begin{equation} \label{neweq12}
   [{\bm \Omega}^{(ij)}]_{rs} = [{\bm \Omega}]_{(i-1)m+r, (j-1)m+s} \, , \quad r,s \in [m].
\end{equation}
Then we have the following equivalence \cite[Sec.\ 2.1]{Kolar2014}
\begin{equation}  \label{neweq24}
  \{ i,j \} \not\in {\cal E} \; \Leftrightarrow \; {\bm \Omega}^{(ij)} = {\bm 0} \, .
\end{equation}

Estimation of differential graphs from multi-attribute data was addressed in \cite{Tugnait2024} in high-dimensional settings using a group-lasso penalty, and related approach of \cite{Zhao2022} also uses a group-lasso penalty (other past work has considered only single-attribute models). Given samples  ${\bm x}(t)$, $t=1,2, \cdots , n_x$, of ${\bm x} = [ {\bm z}_1^\top \; {\bm z}_2^\top \; \cdots \; {\bm z}_p^\top ]^\top \in \mathbb{R}^{mp}$ where ${\bm z}_i \in \mathbb{R}^m$, $i \in [p]$, are jointly Gaussian, and similarly given samples ${\bm y}(t)$, $t=1,2, \cdots , n_y$, of ${\bm y} \in \mathbb{R}^{mp}$, the difference ${\bm \Delta} = {\bm \Omega}_y - {\bm \Omega}_x$ was estimated in \cite{Tugnait2024} to determine the differential graph ${\cal G}_\Delta = \left( V, {\cal E}_\Delta \right)$ with edgeset ${\cal E}_\Delta = \{ \{ k, \ell \} \,:\, \| {\bm \Delta}^{(k \ell)} \|_F \ne 0 \}$. It is well-known that use of non-convex penalties (unlike convex lasso penalty) can yield more accurate results, i.e., they can produce sparse set of solution like lasso, and approximately unbiased coefficients for large coefficients, unlike lasso \cite{Fan2001, Candes2008, Lam2009}. The objective of this paper is to investigate use of non-convex (log-sum \cite{Candes2008} and Smoothly Clipped Absolute Deviation (SCAD) \cite{Fan2001, Lam2009}) penalty functions for estimation of multi-attribute differential graphs. 

\subsection{Related Work}
Non-convex penalties have been extensively used for covariance and graph estimation (see \cite{Lam2009, Zou2008, Tugnait21b, Tugnait22, Wei2023} and references therein) and regression-related problems \cite{Varma2020} (and references therein). However, only \cite{Xu16} and \cite{Tugnait23} have investigated use of non-convex penalties for differential graphs, but only for single-attribute differential graph estimation: SCAD and MCP (minimax concave penalty) in \cite{Xu16} and SCAD and log-sum penalty (LSP) in \cite{Tugnait23}. Counterparts to our Theorems 1 and 2 do not exist in \cite{Tugnait23}. For theoretical analysis,  \cite{Xu16} does not require any irrepresentability condition, unlike this paper. On the other, we do not require a minimum amplitude condition on nonzero elements of ${\bm \Delta}$ but \cite{Xu16} does. Our numerical results show that our LSP-based differential graph estimator significantly outperforms both lasso and SCAD based methods. In \cite{Tugnait2024} differential graphs were estimated using group-lasso (group $\ell_1$) penalty. Here we extend \cite{Tugnait2024} to non-convex penalties. 

As discussed in \cite[Sec.\ I-A]{Tugnait2024}, with the exception of \cite{Tugnait2024}, all prior work on high-dimensional differential graph estimation is focused on single-attribute models. One naive approach would be to estimate the two precision matrices separately by any existing estimator (see \cite{Buhlmann2011, Lam2009, Zou2008, Tugnait21b, Tugnait22, Wei2023} and references therein) and then calculate their difference to estimate the differential graph. (This approach is also applicable to multi-attribute graphs.) This approach estimates twice the number of parameters, hence needs larger sample sizes for same accuracy, and also imposes sparsity constraints on each precision matrix for the methods to work. The same comment applies to methods such as \cite{Danaher2014}, where the two precision matrices and their differences are jointly estimated. If only the difference in the precision matrices is of interest, approaches exist where no sparsity constraints are imposed on individual precision matrices. For instance, direct estimation of the difference in the two precision matrices has been considered for single attribute graphs in \cite{Zhao2014, Xu16, Yuan2017, Jiang2018, Tang2020, Wu2020}, where only the difference is required to be sparse, not the two individual precision matrices. In \cite{Xu16, Yuan2017, Jiang2018, Tang2020, Wu2020} precision difference matrix estimators are based on  a D-trace loss \cite{Zhang2014}, while \cite{Zhao2014} discusses a Dantzig selector type estimator. For more details, we refer the reader to \cite[Sec.\ I-A]{Tugnait2024}.

We consider differential graph estimation without regard to the structure of the original graphs. For instance, there has been considerable interest in modeling the precision matrix as a Laplacian matrix \cite{Kumar2019, Tugnait21b, Medvedovsky2024} where non-convex penalties have been used. The off-diagonal elements of a Laplacian matrix are non-positive. The difference of two Laplacian matrices in not necessarily a Laplacian matrix because the off-diagonal elements of the difference can be positive, negative or zero. Therefore, the methods of \cite{Kumar2019, Tugnait21b, Medvedovsky2024} do not apply to our problem. 

\subsection{Our Contributions}
A penalized D-trace loss function approach for differential graph learning from multi-attribute data was presented in \cite{Tugnait2024} using convex group-lasso penalty. In this paper we extend \cite{Tugnait2024} to non-convex log-sum and SCAD penalties. Two proximal gradient descent methods are presented to optimize the objective function. Theoretical analysis establishing sufficient conditions for consistency in support recovery, convexity and estimation in high-dimensional settings is presented in Theorems 1 and 2. While the non-convex penalized D-trace loss function results in  a non-convex optimization problem, Theorem 2 specifies conditions under which it becomes a convex optimization problem (see Remark 2 in Sec.\ \ref{TA}). These conditions favor log-sum penalty over SCAD. Numerical results based on synthetic and real data are presented to illustrate the proposed approaches. In the synthetic data examples where the ground-truth is known, log-sum penalized D-trace loss significantly outperformed the lasso-penalized D-trace loss as well as SCAD penalized D-trace loss with $F_1$-score and Hamming distance as performance metrics.

\subsection{Outline and Notation} \label{outnot}
The rest of the paper is organized as follows. A penalized D-trace loss function is presented in Sec.\ \ref{GM} for estimation of multi-attribute differential graph using non-convex penalties.  Two proximal gradient descent methods are presented in Sec.\ \ref{SGL} to optimize the non-convex objective function. In Sec.\ \ref{TA} we analyze the properties of the estimator of the difference ${\bm \Delta} = {\bm \Omega}_y - {\bm \Omega}_x$, by following \cite[Theorem 1]{Tugnait2024} pertaining to the lasso penalty. Since the SCAD and log-sum penalties are non-convex, the objective function is  non-convex and in general, any optimization of the objective function will yield only a stationary point. Theorem 1 analyzes the properties of such a stationary point under some sufficient conditions, including an irrepresentability condition (similar condition also used in \cite{Ravikumar2011, Kolar2014, Zhang2014, Yuan2017, Jiang2018, Tugnait2024}). In Theorem 2 we investigate sufficient conditions under which the objective function is strictly convex, thereby ensuring that the stationary point of Theorem 1 is a unique minimum. Numerical results based on synthetic as well as real data are presented in Sec.\ \ref{NE} to illustrate the proposed approach.  Proofs of Theorems 1 and 2 are given in Appendices \ref{append1} and \ref{append2}, respectively.

For a set $V$, $|V|$ or $\mbox{card}(V)$ denotes its cardinality. Given ${\bm A} \in \mathbb{R}^{p \times p}$, we use $\phi_{\min }({\bm A})$, $\phi_{\max }({\bm A})$, $|{\bm A}|$ and $\mbox{tr}({\bm A})$ to denote the minimum eigenvalue, maximum eigenvalue, determinant and  trace of ${\bm A}$, respectively. For ${\bm B} \in \mathbb{R}^{p \times q}$, we define  $\|{\bm B}\| = \sqrt{\phi_{\max }({\bm B}^\top  {\bm B})}$, $\|{\bm B}\|_F = \sqrt{\mbox{tr}({\bm B}^\top  {\bm B})}$, $\|{\bm B}\|_1 = \sum_{i,j} |B_{ij}|$, where $B_{ij}$ is the $(i,j)$-th element of ${\bm B}$ (also denoted by $[{\bm B}]_{ij}$), $\|{\bm B}\|_\infty = \max_{i,j} |B_{ij}|$ and $\|{\bm B}\|_{1,\infty} = \max_i \sum_j|B_{ij}|$. The symbols $\otimes$ and $\boxtimes$ denote Kronecker product and Tracy-Singh product \cite{Tracy1989}, respectively. In particular, given block partitioned matrices ${\bm A} =[{\bm A}_{ij}]$ and ${\bm B}=[{\bm B}_{k \ell}]$  with submatrices ${\bm A}_{ij}$ and ${\bm B}_{k \ell}$, Tracy-Singh product yields another block partitioned matrix ${\bm A} \boxtimes {\bm B} = [{\bm A}_{ij} \boxtimes {\bm B}]_{ij} = [[{\bm A}_{ij} \otimes {\bm B}_{k \ell}]_{k \ell} ]_{ij}$ \cite{Liu2008}. Given ${\bm A} =[{\bm A}_{ij}] \in \mathbb{R}^{mp \times mp}$ with ${\bm A}_{ij} \in \mathbb{R}^{m \times m}$, ${\rm vec}({\bm A}) \in \mathbb{R}^{m^2p^2}$ denotes the vectorization of ${\bm A}$ which stacks the columns of the matrix ${\bm A}$, and 
\begin{align*}
  & {\rm bvec}({\bm A}) =  [({\rm vec}({\bm A}_{11}))^\top \; ({\rm vec}({\bm A}_{21}))^\top \; \cdots 
      \; ({\rm vec}({\bm A}_{p1}))^\top \\
		& \quad ({\rm vec}({\bm A}_{12}))^\top \;  \cdots \; ({\rm vec}({\bm A}_{p2}))^\top \; \cdots \; ({\rm vec}({\bm A}_{pp}))^\top]^\top .
\end{align*} 

Let $S = {\cal E}_\Delta = \{ \{ k, \ell \} \,:\, \| {\bm \Delta}^{(k \ell)} \|_F \ne 0 \}$ where ${\bm \Delta} =[{\bm \Delta}^{(k \ell)}] \in \mathbb{R}^{mp \times mp}$ with ${\bm \Delta}^{(k \ell)} \in \mathbb{R}^{m \times m}$ denoting the $(k,l)$th $m \times m$ submatrix of ${\bm \Delta}$. Then ${\bm \Delta}_S$ denotes the submatrix of ${\bm \Delta}$ with block rows and columns indexed by $S$, i.e., ${\bm \Delta}_S =[{\bm \Delta}^{(k \ell)}]_{(k , \ell)  \in S}$. Suppose ${\bm \Gamma} = {\bm A} \boxtimes {\bm B}$ given block partitioned matrices ${\bm A} =[{\bm A}_{ij}]$ and ${\bm B}=[{\bm B}_{k \ell}]$. For any two subsets  $T_1$ and $T_2$ of  $V \times V$, ${\bm \Gamma}_{T_1,T_2}$ denotes the submatrix of ${\bm \Gamma}$ with block rows and columns indexed by $T_1$ and $T_2$, i.e., ${\bm \Gamma}_{T_1,T_2} = [{\bm A}_{j \ell} \otimes {\bm B}_{kq}]_{(j,k) \in T_1, (\ell,q) \in T_2}$. Following \cite{Kolar2014}, an operator $\bm{\mathcal C}( \cdot )$ is used in Sec.\ \ref{TA}. Consider ${\bm A} \in \mathbb{R}^{mp \times mp}$ with $(k,l)$th $m \times m$ submatrix ${\bm A}^{(k \ell)}$. Then $\bm{\mathcal C}( \cdot )$ operates on ${\bm A}$ as 
\begin{align*}
 \begin{bmatrix} {\bm A}^{(11)} & \cdots & {\bm A}^{(1p)} \\
			  \vdots     & \ddots     & \vdots \\
			{\bm A}^{(p1)} &  \cdots & {\bm A}^{(pp)} \end{bmatrix} \overset{\bm{\mathcal C}( \cdot )}{\xrightarrow{\hspace*{0.5cm}}}
			\begin{bmatrix} \|{\bm A}^{(11)}\|_F & \cdots & \|{\bm A}^{(1p)}\|_F \\
			  \vdots     & \ddots     & \vdots \\
			\|{\bm A}^{(p1)}\|_F &  \cdots & \|{\bm A}^{(pp)}\|_F \end{bmatrix}
\end{align*}
with $\bm{\mathcal C}( {\bm A}^{(k \ell)} ) = \|{\bm A}^{(k \ell)}\|_F$ and $\bm{\mathcal C}( {\bm A} ) \in \mathbb{R}^{p \times p}$. Now consider ${\bm A}_1  , {\bm A}_2 \in \mathbb{R}^{mp \times mp}$ with $(k,l)$th $m \times m$ submatrices ${\bm A}_1^{(k \ell)}$ and ${\bm A}_2^{(k \ell)}$, respectively, and Tracy-Singh product ${\bm A}_1 \boxtimes {\bm A}_2 \in \mathbb{R}^{(mp)^2 \times (mp)^2}$. Then $\bm{\mathcal C}( \cdot )$ operates on ${\bm A}_1 \boxtimes {\bm A}_2$ as $\bm{\mathcal C}({\bm A}_1 \boxtimes {\bm A}_2) \in \mathbb{R}^{p^2 \times p^2}$ with $\bm{\mathcal C}( {\bm A}_1^{(k_1 \ell_1)} \otimes {\bm A}_2^{(k_2 \ell_2)} ) = \|{\bm A}_1^{(k_1 \ell_1)} \otimes {\bm A}_2^{(k_2 \ell_2)} \|_F$ (=$\|{\bm A}_1^{(k_1 \ell_1)}  \|_F \, \|{\bm A}_2^{(k_2 \ell_2)} \|_F$). That is, each $m^2 \times m^2$ submatrix ${\bm A}_1^{(k_1 \ell_1)} \otimes {\bm A}_2^{(k_2 \ell_2)}$ of ${\bm A}_1 \boxtimes {\bm A}_2 $ is mapped into its Frobenius norm.

\section{Penalized D-Trace Loss } \label{GM}
Let ${\bm x} = [ {\bm z}_1^\top \; {\bm z}_2^\top \; \cdots \; {\bm z}_p^\top ]^\top \in \mathbb{R}^{mp}$ where ${\bm z}_i \in \mathbb{R}^m$, $i \in [p]$, are zero-mean, jointly Gaussian. Given i.i.d.\ samples ${\bm x}(t)$, $t=1,2, \cdots , n_x$, of ${\bm x}$, and similarly given i.i.d.\ samples ${\bm y}(t)$, $t=1,2, \cdots , n_y$, of ${\bm y} \in \mathbb{R}^{mp}$, form the sample covariance estimates
\begin{equation}
  \hat{\bm \Sigma}_x = \frac{1}{n_x} \sum_{t=1}^{n_x} {\bm x}(t) {\bm x}^\top(t) \, , \;\;
	\hat{\bm \Sigma}_y = \frac{1}{n_y} \sum_{t=1}^{n_y} {\bm y}(t) {\bm y}^\top(t) \, .  \label{eqn10}
\end{equation} 
and denote their true values as ${\bm \Sigma}_x^\ast = {\bm \Omega}_x^{-\ast} (=({\bm \Omega}_x^\ast)^{-1})$ and ${\bm \Sigma}_y^\ast = {\bm \Omega}_y^{-\ast}$. We wish to estimate ${\bm \Delta} = {\bm \Omega}_y^\ast - {\bm \Omega}_x^\ast$ and graph ${\cal G}_\Delta = \left( V, {\cal E}_\Delta \right)$, based on $\hat{\bm \Sigma}_x$ and $\hat{\bm \Sigma}_y$. Following \cite{Tugnait2024} (see also \cite{Yuan2017}and \cite[Sec.\ 2.1]{Jiang2018}), we use the convex D-trace (difference-in-trace) loss function
\begin{equation}
  L({\bm \Delta}, \hat{\bm \Sigma}_x , \hat{\bm \Sigma}_y) = \frac{1}{2} \mbox{tr} (\hat{\bm \Sigma}_x {\bm \Delta} \hat{\bm \Sigma}_y {\bm \Delta}^\top) 
	   - \mbox{tr} ({\bm \Delta}  (\hat{\bm \Sigma}_x-\hat{\bm \Sigma}_y)) \, . \label{eqn15}
\end{equation}
The  function $L({\bm \Delta}, {\bm \Sigma}_x^\ast , {\bm \Sigma}_y^\ast)$ is strictly convex in ${\bm \Delta}$ and has a unique minimum at ${\bm \Delta}^\ast = {\bm \Omega}_y^\ast - {\bm \Omega}_x^\ast$ \cite{Yuan2017, Jiang2018}. When we use sample covariances, we estimate ${\bm \Delta}$ by minimizing the penalized loss function
\begin{equation}
  L_\lambda({\bm \Delta}, \hat{\bm \Sigma}_x , \hat{\bm \Sigma}_y) = L({\bm \Delta}, \hat{\bm \Sigma}_x , \hat{\bm \Sigma}_y)
	   +  \sum_{k, \ell=1}^p \rho_\lambda \big( \| {\bm \Delta}^{(k \ell)} \|_F  \big) \label{eqn20}
\end{equation}
where, for $u \in \mathbb{R}$, $\rho_\lambda(u)$ is a penalty function that is function of $|u|$, $\lambda > 0$ is a tuning parameter, and $\| {\bm \Delta}^{(k \ell)} \|_F$ promotes blockwise sparsity in ${\bm \Delta}$ \cite{Yuan2006} where, if we partition ${\bm \Delta}$ into $m \times m$ submatrices, ${\bm \Delta}^{(k \ell)}$ denotes its $(k , \ell)$th submatrix, associated with edge $\{ k ,\ell\}$ of edgeset ${\cal E}_{\Delta}$ of the differential graph ${\cal G}_\Delta = \left( V, {\cal E}_\Delta \right)$. {\bf For ease of notation, henceforth, we also use $L_\lambda({\bm \Delta})$, $L({\bm \Delta})$ and $L^\ast({\bm \Delta})$ for $L_\lambda({\bm \Delta}, \hat{\bm \Sigma}_x , \hat{\bm \Sigma}_y)$, $L({\bm \Delta}, \hat{\bm \Sigma}_x , \hat{\bm \Sigma}_y)$ and $L({\bm \Delta}, {\bm \Sigma}_x^\ast , {\bm \Sigma}_y^\ast)$, respectively.}

The following penalty functions are considered: 
\begin{itemize}
\item[(i)] {\it Lasso}. For some $\lambda > 0$,  $\rho_\lambda(u) = \lambda |u|$, $u \in \mathbb{R}$. It is the well-known $\ell_1$-penalty resulting in a convex function that is widely used (also used in \cite{Tugnait2024}). 
\item[(ii)] {\it Log-sum}. For some $\lambda > 0$ and $1 \gg \epsilon > 0$, $\rho_\lambda(u) = \lambda \epsilon \, \ln \big( 1 + \frac{|u|}{\epsilon} \big)$. It is a non-convex function. 
\item[(iii)] {\it Smoothly Clipped Absolute Deviation (SCAD)}. For some $\lambda > 0$ and $a > 2$, $\rho_\lambda(u) = \lambda | u |$ for $|u| \le \lambda$, $= (2 a \lambda | u |- | u |^2 - \lambda^2)/(2 (a-1))$ for $\lambda < |u| < a \lambda$, and $=\lambda^2 (a+1)/2$ for $|u| \ge a$. It is a non-convex function. 
\end{itemize}
In the terminology of \cite{Loh2017}, all of the above three penalties are ``$\mu$-amenable'' for some $\mu \ge 0$. As defined in \cite[Sec.\ 2.2]{Loh2017}, $\rho_\lambda(u)$ is $\mu$-amenable for some $\mu \ge 0$ if the following properties hold: 
\begin{itemize}
\item[(i)] The function $\rho_\lambda(u)$ is symmetric around zero, i.e., $\rho_\lambda(u) = \rho_\lambda(-u)$ and $\rho_\lambda(0) = 0$. 
\item[(ii)] The function $\rho_\lambda(u)$ is non-decreasing on $\mathbb{R}_+$. 
\item[(iii)] The function $\rho_\lambda(u)/u$ is non-increasing on $\mathbb{R}_+$. 
\item[(iv)] The function $\rho_\lambda(u)$ is differentiable for $u \ne 0$. 
\item[(v)] The function $\rho_\lambda(u) +\frac{\mu}{2} u^2$ is convex, for some $\mu \ge 0$. 
\item[(iv)] $\lim_{u \rightarrow 0^+} \rho^\prime(u) = \lambda$ where $\rho^\prime(u) := \frac{d \rho_\lambda (u)}{du}$.  
\end{itemize}
It is shown in \cite[Appendix A.1]{Loh2017}, that all of the above three penalties are $\mu$-amenable with $\mu = 0$ for lasso and $\mu =1/(a-1)$ for SCAD. In \cite{Loh2017} the log-sum penalty is defined as $\rho_\lambda(u) = \ln (1+\lambda |u|)$ whereas in \cite{Candes2008}, it is defined as $\rho_\lambda(u) = \lambda \, \ln \left( 1 + \frac{|u|}{\epsilon} \right)$. We follow  \cite{Candes2008} but modify it so that property (vi) in the definition of $\mu$-amenable penalties holds. In our case $\mu = \frac{\lambda}{\epsilon}$ for the log-sum penalty. 

Suppose
\begin{equation}
  \hat{\bm \Delta} = \arg\min_{\bm \Delta} L_\lambda({\bm \Delta}) \, . \label{eqn25}
\end{equation}
Even though ${\bm \Delta}$ is symmetric, $\hat{\bm \Delta}$ may not be. We can symmetrize it by setting $\hat{\bm \Delta}_{sym} = \frac{1}{2} ( \hat{\bm \Delta} + \hat{\bm \Delta}^\top )$, after obtaining $\hat{\bm \Delta}$.

\section{Optimization} \label{SGL}
The objective function $L_\lambda({\bm \Delta})$ is  non-convex for the non-convex SCAD and log-sum penalties. In this section we discuss two possible optimization approaches to attain a local minimum (actually a stationary point) of $L_\lambda({\bm \Delta})$. Both are based on a proximal gradient descent (PGD) approach which was found to perform better than an alternating direction method of multipliers (ADMM) approach in simulation examples in \cite{Tugnait2024} (with $F_1$-score as the performance metric). 

\subsection{Local Linear Approximation (LLA)} \label{LLA}
Here, for non-convex $\rho_\lambda(u)$, we use a local linear approximation (LLA) to $\rho_\lambda(u)$ as in \cite{Zou2008, Lam2009}, to yield 
\begin{equation}
  \rho_{\lambda}(u) \approx \rho_{\lambda}(|u_0|) 
	 + \rho_\lambda^\prime(|u_0|) (|u| - |u_0|) \, ,
\end{equation}
where $u_0$ is an initial guess, and the gradient of the penalty function is 
\begin{align} 
  \rho_\lambda^\prime(|u_0|) = & \left\{ \begin{array}{l}
       \frac{\lambda \epsilon }{|u_0| +\epsilon}   \mbox{ for log-sum}, \\ 
	 \left\{ \begin{array}{ll} \lambda , & \mbox{if  } |u_0| \le \lambda \\
	   \frac{ a \lambda - | u_0 |}{ a-1} , & \mbox{if  } \lambda < |u_0| \le a \lambda \\ 
	0 , & \mbox{if  } a \lambda < |u_0|  \end{array} \right. \\
		  \quad\quad \mbox{ for SCAD}. \end{array} \right.  
\end{align}
Therefore, with $u_0$ fixed, we need to consider only the term dependent upon $u$ for optimization w.r.t.\ $u$:
\begin{equation}
  \rho_{\lambda}(u)  \, \Rightarrow \, \rho_\lambda^\prime(|u_0|) \, |u| \, .
\end{equation}
By \cite[Theorem 1]{Zou2008}, the LLA provides a majorization of the non-convex penalty, thereby yielding a majorization-minimization approach. 

\begin{algorithm} [t]
\caption{PGD Algorithm under LLA}
\label{alg1}

\algorithmicrequire{\; $\hat{\bm{\Sigma}}_x$, $\hat{\bm{\Sigma}}_y$, tolerance $\delta$, maximum number of iterations $i_{max}$, lasso estimate $\hat{\bm \Delta}_{\rm lasso}$ for SCAD and log-sum penalties} \\
\algorithmicensure{\;\ Estimated $\hat{\bm \Delta}_{sym}$ and $\hat{\cal E}_{\Delta}$.}

\begin{algorithmic}[1] 
\STATE Set $\eta = 1/L_{c1}$ ($L_{c1}$ as in (\ref{eqn108})), ${\bm \Delta}^{(0)} =  {\bm 0}$ for lasso, $ =  \hat{\bm \Delta}_{\rm lasso}$ for SCAD and log-sum penalties.
\STATE converged = FALSE, $i=0$
\WHILE{converged = FALSE $\;$ AND $\;$ $i \le i_{max}$,}
\STATE Set ${\bm A}_1 =  {\bm \Delta}^{(i)} - \eta \nabla {L}({\bm \Delta}^{(i)})$, with $\nabla{L}({\bm \Delta})$ as in (\ref{eqn106}).
\STATE Update $({\bm \Delta}^{(i+1)})^{(k \ell)} =  \big( 1 - \frac{\lambda_{k \ell} \, \eta}
         {  \| {\bm A}_1^{(k \ell)} \|_F } \big)_+     {\bm A}_1^{(k \ell)}$ for $k, \ell \in [p]$.
\STATE  If $\frac{\bar{L}({\bm \Delta}^{(i+1)})-\bar{L}({\bm \Delta}^{(i)})}{\bar{L}({\bm \Delta}^{(i)})} \le \delta$, set converged = TRUE .
\STATE $i \leftarrow i+1$
\ENDWHILE
\STATE Set $\hat{\bm \Delta}_{sym} = \frac{1}{2} ({\bm \Delta} + {\bm \Delta}^\top)$. If $\|\hat{\bm \Delta}_{sym}^{(jk)}\|_F > 0$, assign edge $\{ j,k\} \in \hat{\cal E}_{\Delta}$, else $\{ j,k\} \not\in \hat{\cal E}_{\Delta}$.
\end{algorithmic}
\end{algorithm}

Thus in LSP, with some initial guess $\bar{\bm{\Delta}}$, we replace 
\begin{align}
  \rho_\lambda ( \| {\bm \Delta}^{(k \ell)} \|_F )  & \rightarrow \lambda_{k \ell} :=
	 \frac{\lambda \epsilon }{\| \bar{\bm \Delta}^{(k \ell)} \|_F  +\epsilon} \, .
\end{align}
The solution $\hat{\bm \Delta}_{\rm lasso}$ to the convex lasso-penalized objective function may be used as an initial guess with $\bar{\bm{\Delta}} = \hat{\bm \Delta}_{\rm lasso}$. Similarly, for SCAD, we have 
\begin{align} 
  \lambda_{k \ell} = &
			 \left\{ \begin{array}{ll} \lambda , & \mbox{if  } \| \bar{\bm \Delta}^{(k \ell)} \|_F \le \lambda \\
				    \frac{ a \lambda - \| \bar{\bm \Delta}^{(k \ell)} \|_F}{ a-1} , 
									& \mbox{if  } \lambda < \| \bar{\bm \Delta}^{(k \ell)} \|_F \le a \lambda \\ 
						0 , & \mbox{if  } a \lambda < \| \bar{\bm \Delta}^{(k \ell)} \|_F  \end{array} \right.  \, .  
\end{align}
With LLA, the objective function is transformed to 
\begin{equation}
  \bar{L}_\lambda({\bm \Delta}) 
	      = L({\bm \Delta})
	   +  \sum_{k, \ell=1}^p \lambda_{k \ell} \, \| {\bm \Delta}^{(k \ell)} \|_F \, .   \label{eqn20r3}
\end{equation}
We will use an iterative PGD approach \cite{Beck2009, Boyd2014} to minimize $\bar{L}_\lambda({\bm \Delta})$. It is a first-order method that is based on objective function values and gradient evaluations. It has been used for differential graph estimation with lasso penalty in \cite{Tang2020, Zhao2022, Tugnait2024} where $\lambda_{k \ell} = \lambda$ for all edges $\{k, \ell \}$.  

In the PGD method to minimize $\bar{L}_\lambda({\bm \Delta})$, given the old estimate ${\bm \Delta}^{\rm old}$, in the next iteration, the new estimate ${\bm \Delta}^{\rm new}$ is given by
\begin{align} 
 {\bm \Delta}^{\rm new} = & \arg\min_{\bm \Delta} \Big( \frac{1}{2} \| {\bm \Delta} - {\bm A}_1 \|_F^2 
         + \eta  \, \sum_{k, \ell=1}^p \lambda_{k \ell} \| {\bm \Delta}^{(k \ell)} \|_F \Big)     \label{eqn100}   
\end{align}
where
\begin{align}	
{\bm A}_1 = & {\bm \Delta}^{\rm old} - \eta \nabla {L}({\bm \Delta}^{\rm old}) \, , \label{eqn102}
\end{align}
for a step-size of $\eta$ and $\nabla {L}({\bm \Delta}^{\rm old})$ is the gradient of ${L}({\bm \Delta})$ at the old value. The solution is given by \cite{Yuan2006, Tang2020, Zhao2022}
\begin{align}  
  ({\bm \Delta}^{(k \ell)})^{\rm new}   
	= & \Big( 1 - \frac{\lambda_{k \ell} \eta}{  \| {\bm A}_1^{(k \ell)} \|_F } \Big)_+    
								  {\bm A}_1^{(k \ell)}  , \;\; k, \ell \in [p] \, ,  \label{eqn104}
\end{align}
where $b_+ = \max(0,b)$, $b \in \mathbb{R}$. We have \cite{Yuan2006, Tang2020, Zhao2022}
\begin{align}  
  \nabla{L}({\bm \Delta})   
	= & \hat{\bm \Sigma}_x {\bm \Delta} \hat{\bm \Sigma}_y - (\hat{\bm \Sigma}_x-\hat{\bm \Sigma}_y) \, . \label{eqn106} 
\end{align}
The function ${L}({\bm \Delta})$ is Lipschitz-continuous with Lipschitz constant $L_c$ given by \cite{Yao2018, Tang2020}
\begin{align}  
  L_{c1}   
	= & \phi_{max}(\hat{\bm \Sigma}_x) \, \phi_{max}(\hat{\bm \Sigma}_y) \, .  \label{eqn108}
\end{align}
Therefore, a fixed step-size choice $0 < \eta \le L_{c1}^{-1}$ guarantees convergence of the PGD method to a (local) minimum \cite{Beck2009}, which is a global minimum of the LLA cost since $\bar{L}_\lambda({\bm \Delta})$ is convex. 

A pseudocode for the PGD algorithm is given in Algorithm \ref{alg1}. We minimize $\bar{L}_\lambda({\bm \Delta})$ w.r.t.\ ${\bm \Delta}$ using Algorithm \ref{alg1} as follows: 
\begin{itemize}
\item[(i)] Lasso: $\lambda_{k \ell} = \lambda$ $\forall k, \ell$. 
\item[(ii)] Log-sum: First run lasso to convergence, then use LLA with $\bar{\bm{\Delta}} = \hat{\bm \Delta}_{\rm lasso}$ to obtain $\lambda_{k \ell}$s, and repeat Algorithm \ref{alg1}. 
\item[(iii)]SCAD: Follow the procedure for log-sum but with $\lambda_{k \ell}$s computed for SCAD. 
\end{itemize}

\subsection{Nonconvexity Redistribution} \label{NR}
Let $\tilde{\rho}_\lambda(u) = \rho_\lambda(u) - \lambda |u|$. Then $\tilde{\rho}_\lambda(u)$ is everywhere differentiable with $\tilde{\rho}_\lambda^\prime(0)=0$. Similar to \cite{Loh2017, Yao2018}, we rewrite (\ref{eqn20}) as
\vspace*{-0.1in}
\begin{align}
  L_\lambda({\bm \Delta}) 
	     & = \tilde{L}_\lambda({\bm \Delta})
	   +  \lambda \, \sum_{k, \ell=1}^p \| {\bm \Delta}^{(k \ell)} \|_F    \label{eqn20r1a} 
\end{align}
where
\begin{align}
 \tilde{L}_\lambda({\bm \Delta}) 
    &  = {L}({\bm \Delta})
	   +  \sum_{k, \ell=1}^p \tilde{\rho}_\lambda \big( \| {\bm \Delta}^{(k \ell)} \|_F  \big)  \, ,  \label{eqn20r2a}
\end{align}
$\tilde{L}_\lambda({\bm \Delta})$ is non-convex but smooth and $\sum_{k, \ell=1}^p \| {\bm \Delta}^{(k \ell)} \|_F$ is convex but nonsmooth \cite{Loh2017}. In (\ref{eqn20}), $L({\bm \Delta})$ is convex and smooth but $\rho_\lambda \big( \| {\bm \Delta}^{(k \ell)} \|_F  \big)$ is non-convex and nonsmooth. Now in the PGD approach, given the old estimate ${\bm \Delta}^{\rm old}$, in the next iteration, the new estimate ${\bm \Delta}^{\rm new}$ is given by
\begin{align} 
 {\bm \Delta}^{\rm new} = & \arg\min_{\bm \Delta} \Big( \frac{1}{2} \| {\bm \Delta} - {\bm A}_2 \|_F^2 
         + \eta \lambda \, \sum_{k, \ell=1}^p  \| {\bm \Delta}^{(k \ell)} \|_F \Big)     \label{eqn100a}   
\end{align}
where
\begin{align}	
{\bm A}_2 = & {\bm \Delta}^{\rm old} - \eta \nabla \tilde{L}_\lambda({\bm \Delta}^{\rm old}) \, , \label{eqn102}
\end{align}
for a step-size of $\eta$ and $\nabla \tilde{L}_\lambda({\bm \Delta}^{\rm old})$ is the gradient of $\tilde{L}_\lambda({\bm \Delta})$ at the old value. The solution is given by \cite{Yuan2006, Tang2020, Zhao2022}
\begin{align}  
  ({\bm \Delta}^{(k \ell)})^{\rm new}   
	= & \Big( 1 - \frac{\lambda \eta}{  \| {\bm A}_2^{(k \ell)} \|_F } \Big)_+    
								  {\bm A}_2^{(k \ell)}  , \;\; k, \ell \in [p] \, .  \label{eqn104a}
\end{align}
We have
\begin{align}  
  \nabla\tilde{L}_\lambda({\bm \Delta})   
	= & \hat{\bm \Sigma}_x {\bm \Delta} \hat{\bm \Sigma}_y - (\hat{\bm \Sigma}_x-\hat{\bm \Sigma}_y)   + {\bm G} \, , \label{eqn106a} 
\end{align}
where
\begin{align}
	{\bm G}^{(k \ell)} = & \left\{ \begin{array}{l}
	            {\bm 0}  \mbox{ for lasso} \\
							\lambda \big( \frac{\epsilon}{\epsilon + \|{\bm \Delta}^{(k \ell)}\|_F} -1 \big) 
							          \frac{{\bm \Delta}^{(k \ell)}}{\|{\bm \Delta}^{(k \ell)}\|_F}   \mbox{ for log-sum} \\ 
			 \left\{ \begin{array}{ll} {\bm 0} , & \mbox{if  } \|{\bm \Delta}^{(k \ell)}\|_F \le \lambda \\
				    {\bm B}^{(k \ell)} , 
									& \mbox{if  } \lambda < \|{\bm \Delta}^{(k \ell)}\|_F \le a \lambda \\ 
						-\lambda \frac{{\bm \Delta}^{(k \ell)}}{\|{\bm \Delta}^{(k \ell)}\|_F}, & 
						   \mbox{if  } a \lambda < \|{\bm \Delta}^{(k \ell)}\|_F  \end{array} \right. \\
						  \quad\quad \mbox{ for SCAD} \end{array} \right.  
\end{align}
and \begin{equation}
      {\bm B}^{(k \ell)} = \Big( \frac{a \lambda - \|{\bm \Delta}^{(k \ell)}\|_F}{a-1} 
			          - \lambda \Big) 
             \frac{{\bm \Delta}^{(k \ell)}}{\|{\bm \Delta}^{(k \ell)}\|_F} \, .
\end{equation} 

\begin{algorithm} [t]
\caption{PGD Algorithm after Non-Convexity Redistribution}
\label{alg2}

\algorithmicrequire{\; $\hat{\bm{\Sigma}}_x$, $\hat{\bm{\Sigma}}_y$, tolerance $\delta$, maximum number of iterations $i_{max}$, lasso estimate $\hat{\bm \Delta}_{\rm lasso}$ for SCAD and log-sum penalties} \\
\algorithmicensure{\;\ Estimated $\hat{\bm \Delta}_{sym}$ and $\hat{\cal E}_{\Delta}$.}

\begin{algorithmic}[1] 
\STATE Set $\eta = 1/L_{c2}$ ($L_{c2}$ as in (\ref{eqn108a})), ${\bm \Delta}^{(0)} =  {\bm 0}$ for lasso, $ =  \hat{\bm \Delta}_{\rm lasso}$ for SCAD and log-sum penalties.
\STATE converged = FALSE, $i=0$
\WHILE{converged = FALSE $\;$ AND $\;$ $i \le i_{max}$,}
\STATE Set ${\bm A}_2 =  {\bm \Delta}^{(i)} - \eta \nabla\tilde{L}_\lambda({\bm \Delta}^{(i)})$, with $\nabla\tilde{L}_\lambda({\bm \Delta})$ as in (\ref{eqn106a}).
\STATE Update $({\bm \Delta}^{(i+1)})^{(k \ell)} 
 =  \big( 1 - \frac{\lambda \eta}{  \| {\bm A}_2^{(k \ell)} \|_F } \big)_+    
								  {\bm A}_2^{(k \ell)}$ for $k, \ell \in [p]$.
\STATE  If $\frac{L_\lambda({\bm \Delta}^{(i+1)})-L_\lambda({\bm \Delta}^{(i)})}{L_\lambda({\bm \Delta}^{(i)})} \le \delta$, set converged = TRUE .
\STATE $i \leftarrow i+1$
\ENDWHILE
\STATE Set $\hat{\bm \Delta}_{sym} = \frac{1}{2} ({\bm \Delta} + {\bm \Delta}^\top)$. If $\|\hat{\bm \Delta}_{sym}^{(jk)}\|_F > 0$, assign edge $\{ j,k\} \in \hat{\cal E}_{\Delta}$, else $\{ j,k\} \not\in \hat{\cal E}_{\Delta}$.
\end{algorithmic}
\end{algorithm}

The function $\tilde{L}_\lambda({\bm \Delta})$ is Lipschitz-continuous with Lipschitz constant $L_{c2}$ given by \cite{Yao2018, Tang2020}
\begin{align}  
  L_{c2}   
	= & \left\{ \begin{array}{ll}
	            \phi_{max}(\hat{\bm \Sigma}_x) \, \phi_{max}(\hat{\bm \Sigma}_y) & : \mbox{ Lasso} \\
							\phi_{max}(\hat{\bm \Sigma}_x) \, \phi_{max}(\hat{\bm \Sigma}_y)
							  +  \frac{2m\lambda}{\epsilon} & : \mbox{ Log-sum} \\ 
			 \phi_{max}(\hat{\bm \Sigma}_x) \, \phi_{max}(\hat{\bm \Sigma}_y)
							  + \frac{2m}{a-1} 
						  & : \mbox{ SCAD} \end{array} \right.  \label{eqn108a}
\end{align}
where we have used \cite[Cor.\ 2]{Yao2018} for the log-sum and SCAD penalties. Therefore, a fixed step-size choice $0 < \eta \le L_{c2}^{-1}$ guarantees convergence of the PGD method to a local minimum \cite{Gong2013} provided $L_\lambda({\bm \Delta})$ is bounded below and coercive. We show in Sec.\ \ref{TA} that for large $n$ $(= \min(n_x,n_y))$ this is true with high probability (w.h.p.). A pseudocode for this PGD algorithm is given in Algorithm \ref{alg2}. Since the problem is convex for the lasso penalty, we choose to initialize with the lasso result for log-sum and SCAD.
 
\subsection{Computational Complexity, Convergence and Model Selection} \label{OI}
The computational complexity of the PGD method has been discussed in \cite{Tang2020} for single-attribute differential graphs, and it is of the same order for MA graphs, because the difference lies only in element-wise penalty versus group penalty. Noting that we have $mp \times mp$ precision matrices, by \cite{Tang2020}, the computational complexity of the PGD methods of \cite{Tang2020, Zhao2022} is  either ${\cal O}((mp)^3)$ when as implemented in Algorithms \ref{alg1} and \ref{alg2}, or ${\cal O}((n_x+n_y)(mp)^2)$ when an alternative implementation of the cost gradient in (\ref{eqn106}) is used (see \cite[Sec.\ 2.2]{Tang2020}). For $n_x+n_y \ge mp$, there is no advantage to this alternative approach. 

In the LLA approach, each approximation yields a convex objective function, therefore, convergence to a global minimum is guaranteed. Overall it is a majorization-minimization approach, hence, after repeated LLA's, one gets a local minimum of the original non-convex objective function. In practice, two iterations seem to be enough: first run  Algorithm \ref{alg1} for lasso, then using lasso-based LLA, run Algorithm \ref{alg1} once more. Convergence of the PGD method of Algorithm \ref{alg2} to a local minimum is guaranteed \cite{Gong2013} provided $L_\lambda({\bm \Delta})$ is bounded below and coercive. We show in Sec.\ \ref{TA} that for large $n$ this is true w.h.p. 

For model selection we follow the BIC-like criterion as given in \cite[Sec.\ III-E]{Tugnait2024} (which follows \cite{Yuan2017} who invokes \cite{Zhao2014}):
\begin{align} 
  BIC(\lambda) = & (n_x+n_y) \, \| \hat{\bm \Sigma}_x \hat{\bm \Delta} \hat{\bm \Sigma}_y - (\hat{\bm \Sigma}_x
	   - \hat{\bm \Sigma}_y ) \|_F \nonumber \\
		& \quad + \ln (n_x+n_y) \, | \hat{\bm \Delta} |_0
  \label{eqn1800}  
\end{align}
where $| {\bm A} |_0$ denotes number of nonzero elements in ${\bm A}$ and $\hat{\bm \Delta}$ obeys (\ref{eqn25}). Choose $\lambda$ to minimize $BIC(\lambda)$. Since (\ref{eqn1800}) is not scale invariant, we scale both $\hat{\bm \Sigma}_x$ and $\hat{\bm \Sigma}_y$ (and $\hat{\bm \Delta}$ commensurately) by $\bar{\bm \Sigma}^{-1}$ where $\bar{\bm \Sigma} = \mbox{diag}\{\hat{\bm \Sigma}_x\}$ is a diagonal matrix of diagonal elements of $\hat{\bm \Sigma}_x$.
 
In our simulations we search over $\lambda \in [\lambda_{\ell} , \lambda_{u}]$,  where $\lambda_{\ell}$ and $\lambda_u$ are selected via a heuristic as in \cite{Tugnait21a}. Find the smallest $\lambda$, labeled $\lambda_{sm}$ for which we get a no-edge model; then we set $\lambda_{u}= \lambda_{sm}/2$ and $\lambda_{\ell} = \lambda_{u}/10$. For real data results we picked $\lambda_{u}= \lambda_{sm}$ and $\lambda_{\ell} = \lambda_{u}/5$.

For the numerical results presented later, we picked $i_{\max} = 200$ and $\delta = 10^{-3}$ in Algorithms \ref{alg1} and \ref{alg2}. For the SCAD penalty $a=3.7$ (as in \cite{Lam2009}) and for log-sum penalty $\epsilon = 0.001$.

\section{Theoretical Analysis} \label{TA}  
Here we analyze the properties of $\hat{\bm \Delta}$ specified in (\ref{eqn25}), by following \cite[Theorem 1]{Tugnait2024} pertaining to the lasso penalty. Since the SCAD and log-sum penalties are non-convex, the objective function is  non-convex and in general, any optimization of the objective function will yield only a stationary point. Theorem 1 analyzes the properties of such a stationary point under some sufficient conditions, including an irrepresentability condition (similar condition also used in \cite{Ravikumar2011, Kolar2014, Zhang2014, Yuan2017, Jiang2018, Tugnait2024}). In Theorem 2 we investigate sufficient conditions under which the objective function is strictly convex, thereby ensuring that the stationary point of Theorem 1 is a unique minimum.

Define the true differential edgeset (${\bm \Delta}^{\ast}$ denotes the true value of ${\bm \Delta}$)
\begin{align} 
  S = & {\cal E}_{\Delta^\ast} = \{ \{ k, \ell \} \,:\, \| {\bm \Delta}^{\ast (k \ell)} \|_F \ne 0 \} \, , \quad
	    s = |S| \, .
  \label{eqn200}  
\end{align}
Define $n = \min(n_x,n_y)$, and 
\begin{equation}
  {\bm \Gamma}^\ast = {\bm \Sigma}_y^\ast \boxtimes {\bm \Sigma}_x^\ast \, , \quad
	\hat{\bm \Gamma} = \hat{\bm \Sigma}_y \boxtimes \hat{\bm \Sigma}_x \, . \label{eqn225}
\end{equation}
Also, recall the  operator $\bm{\mathcal C}( \cdot )$ defined in Sec.\ \ref{outnot}.  
In the rest of this section, we allow $p$, $s$ and $\lambda$ to be a functions of sample size $n$, denoted as $p_n$, $s_n$ and $\lambda_n$, respectively.  
Define
\begin{align}
  M & = \max \{ \| \bm{\mathcal C}({\bm \Sigma}_x^\ast) \|_\infty \, , 
	     \| \bm{\mathcal C}({\bm \Sigma}_y^\ast) \|_\infty \} \, , \label{eqn320} \\
	M_\Sigma & = \max \{ \| \bm{\mathcal C}({\bm \Sigma}_x^\ast) \|_{1,\infty} \, , 
	           \| \bm{\mathcal C}({\bm \Sigma}_y^\ast) \|_{1,\infty} \} \, , \label{eqn325} \\
\kappa_\Gamma & = \| \bm{\mathcal C}((\Gamma_{S,S}^\ast)^{-1}) \|_{1,\infty} \, ,\label{eqn330} \\
	\alpha & = 1- \max_{e \in S^c} \| \bm{\mathcal C}({\bm \Gamma}_{e,S}^\ast ({\bm \Gamma}_{S,S}^\ast)^{-1}) \|_{1}  \, ,
	     \label{eqn335} \\
	\bar{\sigma}_{xy} &  = \max \{ \max_i [{\bm \Sigma}_{x}^\ast]_{ii}, \, \max_i [{\bm \Sigma}_{y}^\ast]_{ii} \} \, ,
	     \label{eqn336}  \\
	{C}_0 & = 40 \, m \, \bar{\sigma}_{xy}  \sqrt{2 \big(\tau + \ln(4m^2)/\ln(p_n) \big)}
	     \label{eqn337} 
\end{align}
where $S$ and ${\bm \Gamma}^\ast$ have been defined in (\ref{eqn200}) and (\ref{eqn225}). In (\ref{eqn335}), we require $0 < \alpha < 1$, and the expression 
\[
   \max_{e \in S^c} \| \bm{\mathcal C}({\bm \Gamma}_{e,S}^\ast ({\bm \Gamma}_{S,S}^\ast)^{-1}) \|_{1} \le 1-\alpha
\]
for some $\alpha \in (0,1)$ is called the {\it irrepresentability condition}. Similar conditions are also used in \cite{Ravikumar2011, Kolar2014, Zhang2014, Yuan2017, Jiang2018}.

Let $\partial {L}_\lambda({\bm \Delta})$ denote the sub-differential of ${L}_\lambda({\bm \Delta})$. Recall that we write $\tilde{\rho}_\lambda(u) = \rho_\lambda(u) - \lambda |u|$ so that $\tilde{\rho}_\lambda(u)$ is everywhere differentiable with $\tilde{\rho}_\lambda^\prime(0)=0$. Suppose that $\hat{\bm \Delta}$ is a solution to
\begin{align}
  {\bm 0} \in & \partial {L}_\lambda({\bm \Delta})
	   = \frac{\partial L({\bm \Delta})}{\partial {\bm \Delta}} 
		+ \frac{\partial }{\partial {\bm \Delta}} \Big(
		 \sum_{k, \ell=1}^p \tilde{\rho}_\lambda \big( \| {\bm \Delta}^{(k \ell)} \|_F  \big) \Big) \nonumber \\
		& \quad\quad\quad\quad\quad + \partial  
		 \Big( \lambda \, \sum_{k, \ell=1}^p \| {\bm \Delta}^{(k \ell)} \|_F  \Big) \, , \label{eqn340}
\end{align}
which is a first-order necessary condition for a stationary point of ${L}_\lambda({\bm \Delta})$.  
Theorem 1 addresses some properties of this $\hat{\bm \Delta}$. \\
{\it Theorem 1}. Suppose (\ref{eqn340}) is satisfied for ${\bm \Delta} = \hat{\bm \Delta}$. For the system model of Sec.\ \ref{GM}, under (\ref{eqn200}) and the irrepresentability condition (\ref{eqn335}) for some $\alpha \in (0,1)$, if
\begin{align}  
 &  \lambda_n =  \max \Big\{ \frac{8}{\alpha} \, , \frac{3}{\alpha \bar{C}_\alpha} \, s_n \kappa_\Gamma M  C_{M\kappa} \Big\}
		     {C}_0 \sqrt{\frac{\ln(p_n)}{n}}   \label{eqn350} \\ 
& n = \min(n_x,n_y) >   	\, \max \Big\{\frac{1}{\min\{M^2,1\}} \, , 
              81 M^2 s_n^2 \kappa_\Gamma^2, \nonumber \\
  & \quad\quad \quad\quad \frac{9 s_n^2}{(\alpha \bar{C}_\alpha)^2}  (\kappa_\Gamma M C_{M\kappa})^2 \Big\} 
	  {C}_0^2 \ln(p_n)
		     \label{eqn352}
\end{align}
where $\bar{C}_\alpha = \frac{1-\alpha}{2(2M+1)-2 \alpha M }$ and $C_{M\kappa} = 1.5 \big(1+ \kappa_\Gamma \min\{s_n M^2,M_\Sigma^2\} \big)$, then  $\hat{\bm \Delta}$ is such that with probability $> 1- 2/p_n^{\tau -2}$, for any $\tau >2$, we have
\begin{itemize}
\item[(i)] $\| \bm{\mathcal C}(\hat{\bm \Delta} - {\bm \Delta}^\ast) \|_\infty \le (C_{b1} + C_{b2}) {C}_0 \sqrt{\frac{\ln(p_n)}{n}}$ 
\begin{align*}
 \mbox{where } \;   C_{b1} = & 3 \kappa_\Gamma \, \max \big\{ \frac{8}{\alpha} \, , 
	 \frac{3}{\alpha \bar{C}_\alpha} \, s_n \kappa_\Gamma M C_{M\kappa} \big\} \, , \\
	C_{b2} = & 9 s_n \kappa_\Gamma^2 M^2 \, . 
\end{align*}
\item[(ii)] $\hat{\bm \Delta}_{S^c} = {\bm 0}$.
\item[(iii)] $\| \bm{\mathcal C}(\hat{\bm \Delta} - {\bm \Delta}^\ast) \|_F \le \sqrt{s_n} \, \| \bm{\mathcal C}(\hat{\bm \Delta} - {\bm \Delta}^\ast) \|_\infty$ .
\item[(iv)] Additionally, if $\min_{(k,\ell) \in S} \| ({\bm \Delta}^\ast)^{(k \ell)} \|_F \ge  \newline 2 (C_{b1} + C_{b2}) {C}_0 \sqrt{\frac{\ln(p_n)}{n}}$, then $P({\cal G}_{\hat{\Delta}} = {\cal G}_{{\Delta}^\ast}) > 1- 2/p_n^{\tau -2}$ (support recovery) $\quad \bullet$
\end{itemize}
The proof of Theorem 1 is given in Appendix \ref{append1}.

{{\it Remark 1}. {\it Convergence Rate}}. As discussed in \cite{Tugnait2024}, if $M$, $M_\Sigma$ and $\kappa_\Gamma$ stay bounded with increasing sample size $n$, we have $\| \bm{\mathcal C}(\hat{\bm \Delta} - {\bm \Delta}^\ast) \|_F = {\cal O}_P (s_n^{1.5} \sqrt{ \ln(p_n)/n})$. Therefore, for $\| \bm{\mathcal C}(\hat{\bm \Delta} - {\bm \Delta}^\ast) \|_F \rightarrow 0$ as $n \rightarrow \infty$, we must have $s_n^{1.5} \sqrt{ \ln(p_n)/n} \rightarrow 0$. Notice that $M_\Sigma$ constraints covariances ${\bm \Sigma}_x^\ast$ and ${\bm \Sigma}_y^\ast$ which can be dense even if ${\bm \Omega}_x^\ast$ and ${\bm \Omega}_y^\ast$ are sparse (they need not be sparse for differential estimation), making them possibly unbounded with increasing sample size $n$. In this case we use $\min\{s_n M^2,M_\Sigma^2\}= s_n M^2$ in $C_{M\kappa}$ and $C_{b1}$, with $M$ bounded, leading to $\| \bm{\mathcal C}(\hat{\bm \Delta} - {\bm \Delta}^\ast) \|_F = {\cal O}_P (s_n^{2.5} \sqrt{ \ln(p_n)/n})$.  $\;\; \Box$

Now we vectorize (\ref{eqn15}), using ${\bm \theta} = \mbox{bvec}({\bm \Delta}) \in \mathbb{R}^{m^2p_n^2}$, as (cf.\ (\ref{aeqn220}) in Appendix \ref{append1})
\begin{equation}
	{\cal L}({\bm \theta}) = \frac{1}{2}  {\bm \theta}^\top (\hat{\bm \Sigma}_y \boxtimes  \hat{\bm \Sigma}_x) {\bm \theta}
	  - {\bm \theta}^\top \mbox{bvec}(\hat{\bm \Sigma}_x-\hat{\bm \Sigma}_y) 
	\label{eqn4000}
\end{equation}
where previous $L({\bm \Delta}, \hat{\bm \Sigma}_x , \hat{\bm \Sigma}_y)$ ($=L({\bm \Delta})$) is now ${\cal L}({\bm \theta})$. To include sparse-group penalty, recall that the submatrix ${\bm \Delta}^{(k \ell)}$ of ${\bm \Delta}$ corresponds to the edge $\{k,\ell\}$ of the MA graph. We denote its vectorized version as ${\bm \theta}_{Gt} \in \mathbb{R}^{m^2}$ (subscript $G$ for grouped variables, as in  \cite{Tugnait2024}) with index $t=1,2, \cdots , p_n^2$. Then ${\bm \theta}_{Gt} = \mbox{vec}({\bm \Delta}^{(k \ell)})$ where $t=(k-1)p_n+\ell$, $\ell = t\mod{p_n}$, and $k=\lfloor t/p_n \rfloor +1$.  Using this notation, the penalty $\lambda \sum_{k, \ell=1}^{p_n} \| {\bm \Delta}^{(k \ell)} \|_F = \lambda \sum_{t=1}^{p_n^2} \| {\bm \theta}_{Gt} \|_2$. We now state some restricted strong convexity (RSC) \cite{Negahban2012, Loh2015, Loh2017} results regarding  ${\cal L}({\bm \theta})$ in Lemma 1. Lemma 1(i) deals with the behavior of ${\bm \theta}$ centered on ${\bm \theta}^\ast$ and it implies the RSC in the sense of \cite{Negahban2012}. Lemma 1(ii) deals with the behavior of ${\bm \theta}$ in the subspace consisting of all ${\bm \theta}$'s such that the support of ${\bm \theta}$, $\mbox{supp}({\bm \theta}) \subseteq \mbox{supp}({\bm \theta}^\ast)$, and it implies the RSC in the sense of \cite{Loh2017}.

The proof of Lemma 1 is given in Appendix \ref{append2}. \\
{\it Lemma 1}. (i) Let ${\bm \theta}^\ast = \mbox{bvec}({\bm \Delta}^\ast)$, $\tilde{\bm \theta} = {\bm \theta} - {\bm \theta}^\ast$ and $\phi^\ast_{min} = \phi_{min}({\bm \Sigma}_x^\ast) \phi_{min}({\bm \Sigma}_y^\ast)$. Then with probability $> 1- 2/p_n^{\tau -2}$, for any $\tau >2$ and $\tilde{\bm \theta} \in \mathbb{R}^{m^2p_n^2}$, we have
\begin{align}
      \tilde{\bm \theta}^\top \hat{\bm \Gamma} \tilde{\bm \theta} & 
			  \ge \frac{3}{4} \phi^\ast_{\min} \|\tilde{\bm \theta}\|_2^2
			                  \label{eqn4010}
\end{align} 
if $n > N_2$ where
\begin{equation}
    N_2 = \max \Big\{\frac{1}{M^2} \, , 
              \Big( \frac{192 M s_n}{\phi^\ast_{min}} \Big)^2 \Big\}  {C}_0^2 \ln(p_n) \, .
		     \label{eqn4015}
\end{equation} 
(ii) Let ${\bm \theta}_S = \mbox{bvec}({\bm \Delta}_S) \in \mathbb{R}^{m^2 s_n}$ where ${\bm \Delta}_S$ denotes the submatrix of ${\bm \Delta}$ with block rows and columns indexed by $S$, i.e., ${\bm \Delta}_S =[{\bm \Delta}^{(k \ell)}]_{(k , \ell)  \in S}$, and $s_n = |S|$. Then with probability $> 1- 2/p_n^{\tau -2}$, for any $\tau >2$  we have
\begin{align}
      {\bm \theta}_S^\top \hat{\bm \Gamma}_{S,S} {\bm \theta}_S & \ge \frac{63}{64} \phi^\ast_{\min} \|{\bm \theta}_S\|_2^2
			                  \label{eqn4020}
\end{align} 
for any ${\bm \theta}_S \in \mathbb{R}^{m^2 s_n}$, if $n > N_2$.  $\quad \bullet$

In Lemma 1(i), the support of ${\bm \theta}^\ast = \mbox{bvec}({\bm \Delta}^\ast)$, supp$({\bm \Delta}^\ast)$, is confined to $S$ whereas that of $\tilde{\bm \theta}$ is not. That is, ${\bm \theta}^\ast$ has no more that $m^2 s_n$ nonzero elements whereas none of the elements of $\tilde{\bm \theta}$ need be zero. In Lemma 1(ii), ${\bm \theta}_S$ has only $m^2 s_n$ elements. One may rewrite
\[
  {\bm \theta}_S^\top \hat{\bm \Gamma}_{S,S} {\bm \theta}_S = 
	  {\bm \theta}^\top \hat{\bm \Gamma} {\bm \theta} \, , \quad 
		           \mbox{supp}({\bm \theta}) \subseteq \mbox{supp}({\bm \theta}^\ast) \, .
\]
That is, while $\hat{\bm \Gamma}$ is only positive semi-definite, by (\ref{eqn4020}) $\hat{\bm \Gamma}_{S,S}$ is positive definite. Lemma 1(i) helps in proving that ${\cal L}({\bm \theta})$ (and hence ${\cal L}_\lambda({\bm \theta})$) is bounded from below and coercive, hence ${\cal L}_\lambda({\bm \theta})$ (=$L_\lambda({\bm \Delta})$) has a minimum in the interior \cite[Sec.\ 2.1]{Bertsekas}, so that analyzing (\ref{eqn340}) in Theorem 1 makes sense. Lemma 1(ii) is instrumental to proving Theorem 2, stated in the sequel, where it is shown that $\hat{\bm \Delta}$ of Theorem 1 is a unique minimum of $L_\lambda({\bm \Delta})$.

Setting  ${\bm \theta} = \tilde{\bm \theta}+ {\bm \theta}^\ast$, (\ref{eqn4000}) may be rewritten as 
\begin{align}
	{\cal L}({\bm \theta}) = & \frac{1}{2}  \tilde{\bm \theta}^\top \hat{\bm \Gamma}  \tilde {\bm \theta}
	  + \tilde{\bm \theta}^\top \tilde{\bm b} + {\bm c} 
	\label{eqn4030}
\end{align}
where
\begin{align}
  \tilde{\bm b} = & \hat{\bm \Gamma} {\bm \theta}^\ast + {\bm b} \, ,\\
	{\bm b} = & \mbox{bvec}(\hat{\bm \Sigma}_x-\hat{\bm \Sigma}_y) \, , \\
	{\bm c} = & \frac{1}{2}  {\bm \theta}^{\ast \top} \hat{\bm \Gamma}   {\bm \theta}^\ast
	  - {\bm \theta}^{\ast \top} {\bm b} \, .
\end{align}
It then follows that
\begin{align}
	{\cal L}({\bm \theta}) \ge & \frac{3}{8} \phi^\ast_{\min} \|\tilde {\bm \theta}\|_2^2 
	           - \|\tilde{\bm b}\|_2 \|\tilde {\bm \theta}\|_2 - \|{\bm c}\|_2 \label{eqn4035} \\
			\ge & -\frac{2}{3 \phi^\ast_{\min}} \|\tilde{\bm b}\|_2 - \|{\bm c}\|_2  \, ,
	\label{eqn4040}
\end{align}
implying that ${\cal L}({\bm \theta})$ ($=L({\bm \Delta})$) is bounded from below. Since the penalty $\rho_\lambda(u) \ge 0$, ${\cal L}_\lambda({\bm \theta})$ ($=L_\lambda({\bm \Delta})$) is also bounded from below. By (\ref{eqn4035}), $\lim_{\|\tilde{\bm \theta}\|_2 \rightarrow \infty} {\cal L}({\bm \theta}) \rightarrow \infty$, and since $\|\tilde{\bm \theta}\|_2 \le \|{\bm \theta}\|_2 + \|{\bm \theta}^\ast\|_2$, we have $\lim_{\|{\bm \theta}\|_2 \rightarrow \infty} {\cal L}({\bm \theta}) \rightarrow \infty$, hence coercive. Since $\rho_\lambda(u)$ is non-decreasing on $\mathbb{R}_+$,  ${\cal L}_\lambda({\bm \theta})$) is bounded from below and coercive, hence ${\cal L}_\lambda({\bm \theta})$ (=$L_\lambda({\bm \Delta})$) has a minimum in the interior \cite[Sec.\ 2.1]{Bertsekas}.

Theorem 2 is proved in Appendix \ref{append2}. \\
{\it Theorem 2}. Under Theorem 1, if $n > N_2$ and 
\begin{align}  
	&	\phi_{\min}({\bm \Sigma}^\ast_y) \phi_{\min}({\bm \Sigma}^\ast_x) >  \left\{ \begin{array}{ll}
		   0 & : \;\; \mbox{lasso} \\
			  \frac{64}{63} \times \frac{1}{a-1} & : \;\; \mbox{SCAD} \\
				 \frac{64  }{63} \times \frac{ \lambda_n }{\epsilon} & : \;\; \mbox{log-sum}, \end{array} \right. \label{eqn810} 
\end{align}
then with probability $> 1- 2/p_n^{\tau -2}$, $\tau >2$, $\hat{\bm \Delta}$ of Theorem 1 is a unique minimizer of $L_\lambda({\bm \Delta})$. $\quad \bullet$

{\it Remark 2}. The proof of Theorem 2 relies on the fact that under (\ref{eqn810}), $L({\bm \Delta}_S)$ (=${\cal L}({\bm \theta}_S)$) is strictly convex in ${\bm \Delta}_S$ (equivalently, in ${\bm \theta}_S$). We see that as $n \rightarrow \infty$, $\lambda_n \rightarrow 0$ (see also Remark 1), therefore, we eventually have convexity for log-sum penalty by (\ref{eqn810}) regardless of the value of $\phi_{\min}({\bm \Sigma}^\ast_y) \phi_{\min}({\bm \Sigma}^\ast_x)$ (assuming the latter does not change with $p_n$). But such is not necessarily the case for SCAD.  For SCAD one may need $a$ to become large in which case it would behave more like lasso. $\quad \Box$

{\it Remark 3}. Here we compare our theoretical results Theorems 1 and 2 with Theorem 1 of \cite{Tugnait2024}. Theorem 1 of this paper is analogous to \cite[Theorem 1]{Tugnait2024} where the latter was established for the global optimum of the lasso penalized objective function, whereas Theorem 1 of this paper holds for any stationary point of our non-convex penalized objective function ${L}_\lambda({\bm \Delta})$. The novelty lies in the proof modifications to handle non-convex penalties. Theorem 2 establishes conditions under which the stationary point analyzed in Theorem 1 of this paper is indeed a unique minimizer of  ${L}_\lambda({\bm \Delta})$ (this result is not needed in \cite{Tugnait2024} as the penalized objective function is convex). $\quad \Box$

\begin{table*}
\vspace*{-0.1in}
\caption{{\it ER Graph: $F_1$ scores, Hamming distances, normalized Frobenius norm of estimation error ($\| \hat{\bm \Delta} - {\bm \Delta}^\ast \|_F / \| {\bm \Delta}^\ast \|_F$) and timing, for the synthetic data example ($p=100$, $m=4$), averaged over 100 runs (standard deviation $\sigma$ in parentheses). The BIC method is from \cite[Sec.\ III-E]{Tugnait2024}.}} \label{table1} 
\vspace*{-0.15in}
\begin{center}
\begin{tabular}{ccccc}   \hline\hline
 $n$ &  200 &  400  & 800 & 1600 \\  \hline\hline
\multicolumn{5}{c}{$F_1$ score ($\sigma$): $\lambda$'s picked to maximize $F_1$ } \\ \hline
Lasso  &  0.549 (0.099)  &  0.732 (0.078)  & 0.865 (0.063) & 0.963  (0.036)    \\ 
Log-sum   &  0.591 (0.072)  &  0.759 (0.073)  & 0.908 (0.050) & 0.981 (0.017)     \\ 
	SCAD   &  0.436 (0.063)  &  0.642 (0.066)  & 0.831 (0.066) & 0.956  (0.040)    \\ \hline\hline
\multicolumn{5}{c}{Hamming distance ($\sigma$): $\lambda$'s picked to maximize $F_1$ } \\ \hline
Lasso  &  189.2 (32.9)  &  130.7 (33.0)  & 64.1 (29.9) & 18.5  (17.6)   \\ 
Log-sum   &  203.8 (41.5) &  120.7 (32.6) & 44.3  (23.8) & 09.5 (08.5) \\ 
SCAD   &  290.2 (44.2) &  193.7 (36.2) & 81.5 (32.1) & 21.6 (19.2)   \\ \hline\hline
\multicolumn{5}{c}{Est.\ error ($\sigma$): $\lambda$'s picked to maximize $F_1$ } \\ \hline
Lasso  &  0.954 (0.022)  &  0.895 (0.035)  & 0.856 (0.043) & 0.764   (0.054)   \\ 
Log-sum   &  0.927 (0.037)   &  0.844 (0.045)  & 0.762 (0.057) & 0.587 (0.073)     \\ 
	SCAD   &  1.055 (0.049)  &  0.879 (0.049)  & 0.714 (0.070) & 0.552  (0.050)     \\ \hline\hline
\multicolumn{5}{c}{Timing (s) ($\sigma$): $\lambda$'s picked to maximize $F_1$ } \\ \hline
Lasso  &  06.44 (0.033)  &  06.62 (0.583)  & 06.50 (0.258) & 06.60   (0.355)   \\ 
Log-sum   &  08.26 (0.690)  &  07.42 (0.673)  & 07.19 (1.139) & 10.94   (1.607)     \\ 
	SCAD   &  16.45 (0.096)  &  16.84 (1.007)  & 16.55 (0.538) & 16.81  (0.695)     \\ \hline\hline
\multicolumn{5}{c}{$F_1$ score $\sigma$): $\lambda$'s picked to minimize BIC } \\ \hline
Log-sum   &  0.349 (0.099)  &  0.699 (0.074) & 0.880 (0.100) & 0.980  (0.065)    \\  \hline\hline
\multicolumn{5}{c}{Hamming distance $\sigma$): $\lambda$'s picked to minimize BIC } \\ \hline
Log-sum   &  990.5 (500.4)  &  200.4 (97.3) & 52.7 (37.8) & 8.3 (23.5)      \\ \hline\hline
\end{tabular}  
\end{center}
\end{table*} 

\begin{table*}
\vspace*{-0.1in}
\caption{{\it BA Graph: $F_1$ scores, Hamming distances and normalized Frobenius norm of estimation error ($\| \hat{\bm \Delta} - {\bm \Delta}^\ast \|_F / \| {\bm \Delta}^\ast \|_F$)  and timing, for the synthetic data example ($p=100$, $m=4$), averaged over 100 runs (standard deviation $\sigma$ in parentheses). The BIC method is from \cite[Sec.\ III-E]{Tugnait2024}.}} \label{table2} 
\vspace*{-0.15in}
\begin{center}
\begin{tabular}{ccccc}   \hline\hline
 $n$ &  200 &  400  & 800 & 1600 \\  \hline\hline
\multicolumn{5}{c}{$F_1$ score ($\sigma$): $\lambda$'s picked to maximize $F_1$ } \\ \hline
Lasso  &  0.665 (0.082)  &  0.776 (0.050)  & 0.859 (0.071) & 0.949  (0.034)    \\ 
Log-sum   &  0.693 (0.078)  &  0.808 (0.049)  & 0.896 (0.046) & 0.989 (0.011)     \\ 
	SCAD   &  0.555 (0.069)  &  0.682 (0.051)  & 0.806 (0.071) & 0.932  (0.038)    \\ \hline\hline
\multicolumn{5}{c}{Hamming distance ($\sigma$): $\lambda$'s picked to maximize $F_1$ } \\ \hline
Lasso  &  163.0 (34.7)  &  125.7 (33.6)  & 67.2 (31.5) & 24.7  (15.9)   \\ 
Log-sum   &  150.4 (33.4) &  101.5 (27.1) & 57.0  (28.8) & 05.7 (05.3) \\ 
SCAD   &  236.6 (34.5) &  196.4 (46.5) & 96.1 (32.2) & 33.4 (18.2)   \\ \hline\hline
\multicolumn{5}{c}{Est.\ error ($\sigma$): $\lambda$'s picked to maximize $F_1$ } \\ \hline
Lasso  &  0.955 (0.033)  &  0.834 (0.041)  & 0.840 (0.050) & 0.774   (0.055)   \\ 
Log-sum   &  0.937 (0.047)   &  0.787 (0.049)  & 0.647 (0.057) & 0.466 (0.046)     \\ 
	SCAD   &  0.953 (0.049)  &  0.775 (0.046)  & 0.716 (0.074) & 0.593  (0.069)     \\ \hline\hline
\multicolumn{5}{c}{Timing (s) ($\sigma$): $\lambda$'s picked to maximize $F_1$ } \\ \hline
Lasso  &  07.32 (0.185)   &  07.29 (0.138)  & 07.22 (0.205) & 06.27 (0.434)   \\ 
Log-sum   &  08.45 (0.323)   &  07.90 (0.204)  & 09.65 (2.716) & 10.98 (0.381)     \\ 
	SCAD   &  18.33 (0.305)  &  18.28 (0.250)  & 18.22 (0.304) & 17.27  (0.483)     \\ \hline\hline
\multicolumn{5}{c}{$F_1$ score $\sigma$): $\lambda$'s picked to minimize BIC } \\ \hline
Log-sum   &  0.455 (0.078)  &  0.732 (0.065) & 0.852 (0.090) & 0.917  (0.139)    \\  \hline\hline
\multicolumn{5}{c}{Hamming distance $\sigma$): $\lambda$'s picked to minimize BIC } \\ \hline
Log-sum   &  524.1 (238.9)  &  148.3 (48.7) & 64.0 (34.5) & 31.7 (48.8)      \\ \hline\hline
\end{tabular}  
\end{center}
\end{table*}

\section{Numerical Examples} \label{NE}
We now present numerical results for synthetic as well as real data to illustrate the proposed non-convex penalty approaches. In the synthetic data examples the ground truth is known and this allows for an assessment of the efficacy of various approaches. In the real data example the ground truth is unknown and our goal there is visualization and exploration of the differential conditional dependency structures underlying the data. As noted in Sec.\ \ref{OI}, for all numerical results we picked $i_{\max} = 200$ and $\delta = 10^{-3}$ in Algorithms \ref{alg1} and \ref{alg2}. For the SCAD penalty $a=3.7$ (as in \cite{Lam2009, Varma2020}) and for log-sum penalty $\epsilon = 0.001$. Only $\lambda$ was treated as a tuning parameter.

\subsection{Synthetic Data: Erd\"{o}s-R\`{e}nyi and Barab\'{a}si-Albert Graphs} \label{NEsyn}
We consider two types of graphs: Erd\"{o}s-R\`{e}nyi (ER) graph and Barab\'{a}si-Albert (BA) graph \cite{Barabasi1999, Lu2014}.  In the ER graph, $p=100$ nodes are connected to each other with probability $p_{er} =0.5$ and there are $m=4$ attributes per node whereas in the BA graph, we used $p=100$ and mean degree of 2 to generate a BA graph using the procedure given in \cite{Lu2014}. In the upper triangular $\bm{\Omega}_x$, we set $[\bm{\Omega}_x^{(jk)}]_{st} = 0.5^{|s-t|}$ for $j=k=1, \cdots, p$, $s,t=1, \cdots , m$. For $j \ne k$, if the two nodes are not connected in the graph (ER or BA), we have  $\bm{\Omega}^{(jk)} = {\bm 0}$, and if nodes $j$ and $k$ are connected, then $[\bm{\Omega}^{(jk)}]_{st}$ is uniformly distributed over $[-0.4,-0.1] \cup [0.1,0.4]$ for $s \ne t$, otherwise it is zero. Then add lower triangular elements to make $\bm{\Omega}_x$ a symmetric matrix. To generate $\bm{\Omega}_y$, we follow \cite{Yuan2017} and first generate a differential graph with ${\bm \Delta} \in \mathbb{R}^{(mp) \times (mp)}$ as an ER graph (regardless of whether $\bm{\Omega}_x$ is based on ER or BA model), with connection probability $p_{er} =0.05$ (sparse): if nodes $j$ and $k$ are connected in the $\bm{\Omega}_x$ model, then each of $m^2$ elements of $\bm{\Delta}^{(jk)}$ is independently set to $\pm 0.9$ with equal probabilities. Then $\bm{\Omega}_y = \bm{\Omega}_x + {\bm \Delta}$. Finally add $\gamma {\bm I}$ to  $\bm{\Omega}_y$ and to $ \bm{\Omega}_x$ and pick $\gamma$ so that $\bm{\Omega}_y$ and $\bm{\Omega}_x$ are both positive definite.  With $\bm{\Phi}_x \bm{\Phi}_x^\top =\bm{\Omega}_x^{-1}$, we generate ${\bm x} = \bm{\Phi} {\bm w}$ with ${\bm w} \in \mathbb{R}^{mp}$ as zero-mean Gaussian, with identity covariance, and similarly for ${\bm y}$. We generate $n=n_x=n_y$ i.i.d.\ observations for ${\bm x}$ and ${\bm y}$, with $m=4$, $p =100$, $n \in \{200, 400, 800, 1600\}$.

Simulation results based on 100 runs are shown in Tables \ref{table1} and \ref{table2} for ER and BA graphs, respectively, where the performance measure are $F_1$-score and Hamming distance (between estimated and true edgesets $\hat{\cal E}_{\bm \Delta}$ and ${\cal E}_{{\bm \Delta}^\ast}$) for efficacy in edge detection, normalized estimation error $\| \hat{\bm \Delta} - {\bm \Delta}^\ast \|_F / \| {\bm \Delta}^\ast \|_F$ and execution time (based on tic-toc functions in MATLAB). All algorithms were run on a Window 10 Pro operating system with processor Intel(R) Core(TM) i7-10700 CPU @2.90 GHz with 32 GB RAM, using MATLAB R2023a. For lasso and SCAD ($a$=3.7) penalties we used Algorithm \ref{alg2} whereas for log-sum penalty ($\epsilon = 0.001$) we used LLA-based PGD approach of Algorithm \ref{alg1} since Algorithm \ref{alg2} did not work for log-sum penalty as the Lipschitz constant $L_c$ is too high resulting in extremely small step-sizes offering little improvement over lasso. For SCAD, Algorithm \ref{alg2} yielded better results compared to LLA-based PGD method of Algorithm \ref{alg1}. It is seen that log-sum penalty outperforms lasso and SCAD with $F_1$ score as the performance metric. For $n \ge 800$, SCAD yields smaller estimation errors in estimating ${\bm \Delta}$ for ER graphs in Table \ref{table1} but its performance in terms of $F_1$ score, Hamming distance and execution time metrics is, in general, poor. In practice we do not know the ground truth, hence cannot pick $\lambda$ to maximize the $F_1$ score. In Tables \ref{table1} and \ref{table2} we also show results for log-sum penalty when $\lambda$ is picked based on a BIC method given in \cite[Sec.\ III-E]{Tugnait2024} and discussed in Sec.\ \ref{OI}.

\begin{figure*}
\begin{subfigure}[b]{0.5\textwidth}
\begin{center}
\includegraphics[width=0.8\linewidth]{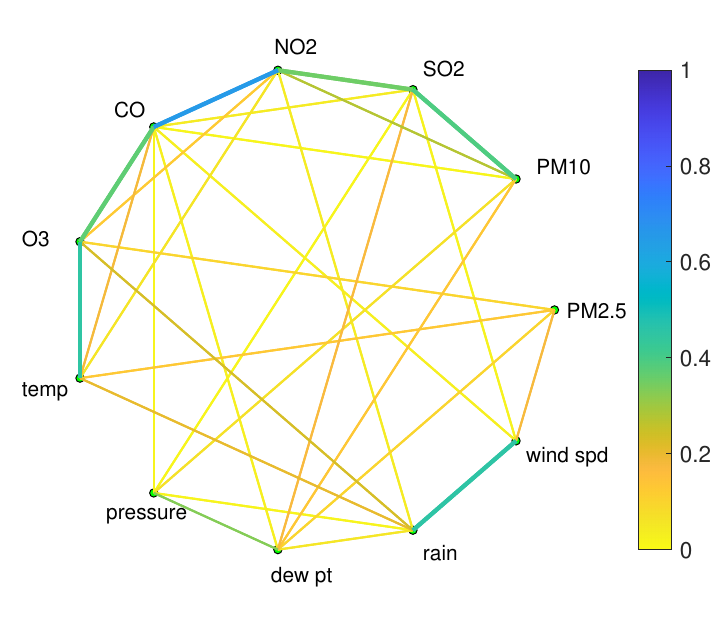}
\caption{Lasso penalty: edges $\big\{ \{k, \ell\} \, : \, \| \hat{\bm \Delta}^{(k \ell)} \|_F > 0 \big\}$}
\end{center}
\end{subfigure}%
\begin{subfigure}[b]{0.5\textwidth}
\begin{center}
\includegraphics[width=0.8\linewidth]{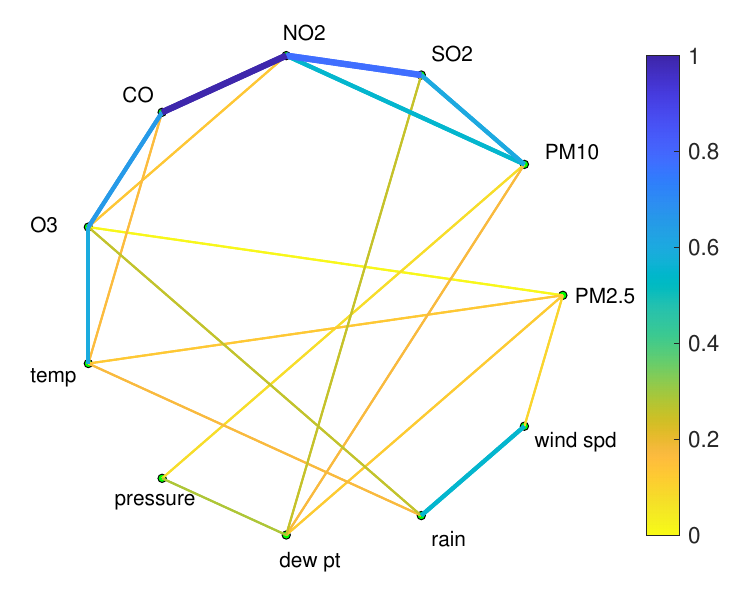}
\caption{Log-sum penalty: edges $\big\{ \{k, \ell\} \, : \, \| \hat{\bm \Delta}^{(k \ell)} \|_F > 0 \big\}$}
\end{center}
\end{subfigure}%
\newline
\begin{subfigure}[b]{0.5\textwidth}
\begin{center}
\includegraphics[width=0.8\linewidth]{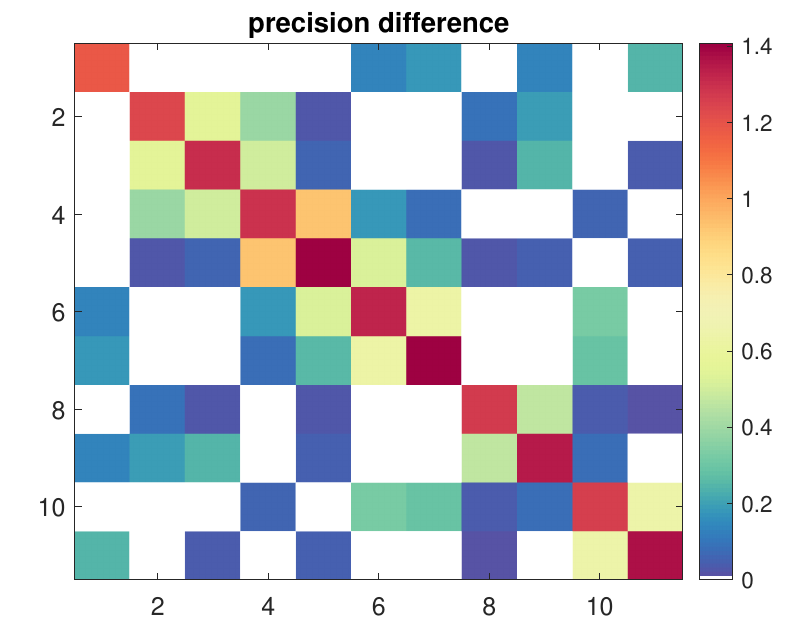}
\caption{Lasso penalty: $\| \hat{\bm \Delta}^{(k \ell)} \|_F$, $k, \ell \in [11]$}
\end{center}
\end{subfigure}%
\begin{subfigure}[b]{0.5\textwidth}
\begin{center}
\includegraphics[width=0.8\linewidth]{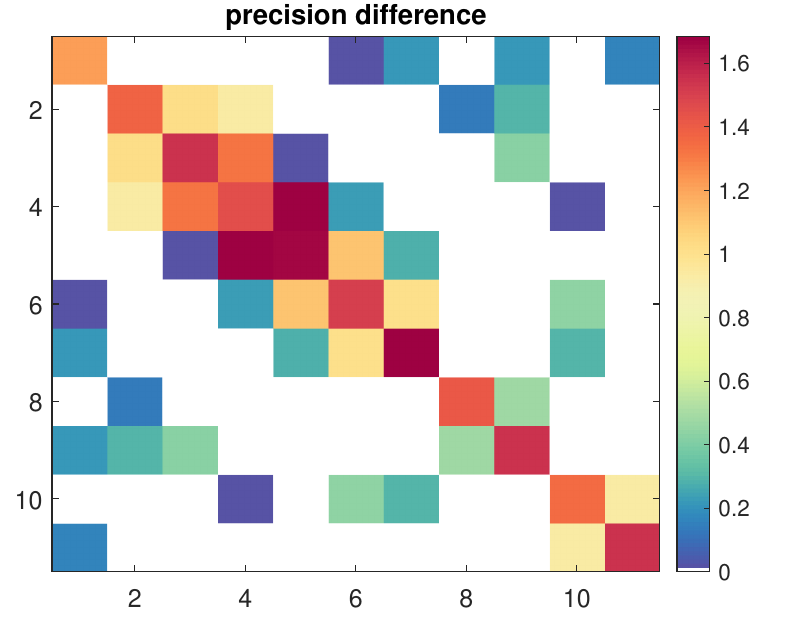}
\caption{Log-sum penalty: $\| \hat{\bm \Delta}^{(k \ell)} \|_F$, $k, \ell \in [11]$}
\end{center}
\end{subfigure}%
\vspace*{-0.05in}
\caption{\small{Differential graphs comparing Beijing air-quality datasets \cite{Zhang2017} acquired from two sets of monitoring stations, 4 stations per set, year 2013-14: 4 monitoring stations and 11 features ($m=4$, $p=11$, $n_x=n_y=365$). Number of distinct edges $=35$ and $21$ in graphs (a) and (b), respectively. Estimated $\|\hat{\bm \Delta}^{(ij)}\|_F$ is the edge weight (normalized to have $\max_{i, j}\|\hat{\bm \Delta}^{(ij)}\|_F=1$). The edge weights are color coded , in addition to the edges with higher weights being drawn thicker.}} \label{figreal}
\vspace*{-0.2in}
\end{figure*}

\subsection{Real Data: Beijing air-quality dataset} \label{NEreal}
Here we consider Beijing air-quality dataset \cite{Zhang2017, Chen2015}, downloaded from \url{https://archive.ics.uci.edu/ml/datasets/Beijing+Multi-Site+Air-Quality+Data}. This data set includes hourly air pollutants data from 12 nationally-controlled air-quality monitoring sites in the Beijing area from the Beijing Municipal Environmental Monitoring Center, and meteorological data in each air-quality site are matched with the nearest weather station from the China Meteorological Administration. The time period is from March 1st, 2013 to February 28th, 2017. The six air pollutants are PM$_{2.5}$, PM$_{10}$, SO$_2$, NO$_2$, CO, and O$_3$, and the meteorological data is comprised of five features: temperature, atmospheric pressure, dew point, wind speed, and rain; we did not use wind direction. Thus we have eleven features. We used data from 8 sites: 4 suburban/rural sites -- Changping, Huairou, Shunyi, Dingling,  and 4 urban area stations -- Aotizhongxin, Dongsi, Guanyuan, Gucheng \cite[Fig.\ 1]{Chen2015}. The data are averaged over 24 hour period to yield daily averages. We used one year of daily data resulting in $n_x = n_y = 365$ days. The stations are used as attributes, with $m=4$, for comparison between suburban/rural sites and urban sites using 2013-14 year data.

We pre-process the data as follows. Given $i$th feature data ${\bm z}_i(t) \in \mathbb{R}^m$, we transform it to $\bar{\bm z}_i(t) = \ln({\bm z}_i(t)/{\bm z}_i(t-1))$ and then detrend it (i.e., remove the best straight-line fit using the MATLAB function detrend). Finally, we scale the detrended scalar sequence to have a mean-square value of one over $n_x$ or $n_y$ samples. The logarithmic transformation and detrending of each feature sequence makes the sequence closer to (univariate) stationary and Gaussian, while scaling ``balances'' the possible wide variations in the scale of various feature measurements. All temperatures were converted from Celsius to Kelvin to avoid negative numbers, and if a value of a feature is zero (e.g., wind speed), we added a small positive number to it, so that the logarithmic transformation is well-defined.

Fig.\ \ref{figreal} shows the estimated differential graphs when comparing daily-averaged data over the period 2013-14, from four suburban/rural sites ($x$-data) to that from four urban sites ($y$-data), with air-quality and meteorological variables as $p=11$ features measured at two sets of 4 monitoring sites ($m$=4). We used the PGD approach of Algorithm \ref{alg1} for log-sum penalty. Model selection was done as in \cite[Sec.\ III-E]{Tugnait2024} and in Sec.\ \ref{OI}. The objective is to visualize and explore differential conditional dependency relationships among the 11 variables, comparing one subregion to another. There are significant differences in meteorological conditions and pollutant sources, levels and mutual interactions, among suburban and urban areas \cite{Zhang2017, Chen2015}. The suburban areas (located toward north) are less polluted than the urban areas (located toward south) \cite{Zhang2017, Chen2015}. Automobile exhaust is the main cause of NO$_2$ which is likely to undergo a chemical reaction with Ozone O$_3$, thereby, lowering its concentration \cite{Chen2015}. Cold, dry air from the north reduces both dew point and PM$_{2.5}$ particle concentration in suburban areas while southerly wind brings warmer and more humid air from the more polluted south that elevates the PM$_{2.5}$ concentration \cite{Zhang2017}. The urban stations neighbor the south of Beijing which is heavily installed with iron, steel and cement industries in Hebei province \cite{Zhang2017}. 

Figs.\ \ref{figreal}(a)-(d) show estimated $\| \hat{\bm \Delta}^{(k \ell)} \|_F$ for various edges $\{k, \ell \}$, where it is unscaled in Fig.\ \ref{figreal}(c),(d) but scaled in Fig.\ \ref{figreal}(a),(b) so that the largest $\| \hat{\bm \Delta}^{(k \ell)} \|_F$ (including $k=\ell$) is normalized to one. It is seen that the non-convex log-sum penalty yields a much sparser differential graph.

\section{Conclusions}
A penalized D-trace loss function approach for differential graph learning from multi-attribute data was presented in \cite{Tugnait2024} using convex group-lasso penalty. In this paper we extended \cite{Tugnait2024} to non-convex log-sum and SCAD penalties. Two proximal gradient descent methods were presented to optimize the objective function. Theoretical analysis establishing sufficient conditions for consistency in support recovery, convexity and estimation in high-dimensional settings was provided in Theorems 1 and 2. Numerical results based on synthetic and real data were presented to illustrate the proposed approaches. In the synthetic data examples the log-sum penalized D-trace loss significantly outperformed the lasso-penalized D-trace loss as well as SCAD penalized D-trace loss with $F_1$-score and Hamming distance as performance metrics.

\appendices
\section{Technical Lemmas and Proof of Theorem 1} \label{append1}
In this Appendix, we provide a proof of Theorem 1. A first-order necessary condition for minimization of non-convex $L_\lambda({\bm \Delta}, \hat{\bm \Sigma}_x , \hat{\bm \Sigma}_y)$ (=$L_\lambda({\bm \Delta}))$, given by (\ref{eqn20}), w.r.t.\ ${\bm \Delta} \in \mathbb{R}^{mp \times mp}$ is that the zero matrix belongs to the sub-differential of $L_\lambda({\bm \Delta})$ at the solution $\hat{\bm \Delta}$. That is,
\begin{align} 
  {\bm 0} = & \frac{\partial L({\bm \Delta})}{\partial {\bm \Delta}}
	     + \lambda {\bm Z}({\bm \Delta}) \, \Big|_{{\bm \Delta} = \hat{\bm \Delta}} \nonumber \\
			= & \hat{\bm \Sigma}_x \hat{\bm \Delta} \hat{\bm \Sigma}_y - (\hat{\bm \Sigma}_x-\hat{\bm \Sigma}_y) 
			           + \lambda {\bm Z}(\hat{\bm \Delta})
  \label{aeqn210}  
\end{align}
where $\lambda {\bm Z}({\bm \Delta}) \in \partial \sum_{k, \ell=1}^p \rho_\lambda \big( \| {\bm \Delta}^{(k \ell)} \|_F \big) \in \mathbb{R}^{mp \times mp}$, the sub-differential of (possibly non-convex) penalty term, is given by 
\begin{align} 
  ({\bm Z}({\bm \Delta}))^{(k \ell)} =& \left\{ \begin{array}{l} 
	  {\bm V} \in \mathbb{R}^{m \times m} , \, \| {\bm V} \|_F \le 1, \\
		\quad\quad \mbox{ if } 
				                  \| {\bm \Delta}^{(k \ell)} \|_F = 0 \\
	      \frac{{\bm \Delta}^{(k \ell)}}{\| {\bm \Delta}^{(k \ell)} \|_F} \mbox{ if } 
				                  \| {\bm \Delta}^{(k \ell)} \|_F \ne 0 \; : \mbox{ lasso}\\
		C^{(k \ell)} \mbox{ if } 
				                  \| {\bm \Delta}^{(k \ell)} \|_F \ne 0	\; : \mbox{ log-sum}	\\	
		D^{(k \ell)} \mbox{ if } 
				                  \| {\bm \Delta}^{(k \ell)} \|_F \ne 0	\; : \mbox{ SCAD} \, ,	\end{array} \right. \label{aeqn215} 
\end{align}
\begin{align} 
		C^{(k \ell)} = & \frac{\epsilon}{\epsilon + \| {\bm \Delta}^{(k \ell)} \|_F} \, 
	      \frac{{\bm \Delta}^{(k \ell)}}{\| {\bm \Delta}^{(k \ell)} \|_F} \, ,
\end{align}
\begin{align} 
		D^{(k \ell)} = & \left\{ \begin{array}{l} 
		\frac{{\bm \Delta}^{(k \ell)}}{\| {\bm \Delta}^{(k \ell)} \|_F} \mbox{ if } 
				                  0 < \| {\bm \Delta}^{(k \ell)} \|_F \le \lambda \\
		 \frac{a - \| {\bm \Delta}^{(k \ell)} \|_F/ \lambda}{a-1} 
		  \frac{{\bm \Delta}^{(k \ell)}}{\|{\bm \Delta}^{(k \ell)}\|_F} \\
			\quad\quad \mbox{ if } 
				                  \lambda < \| {\bm \Delta}^{(k \ell)} \|_F \le a \lambda \\
							{\bm 0} \mbox{ if } 
				                 a \lambda < \| {\bm \Delta}^{(k \ell)} \|_F \, . \end{array} \right.
\end{align}
We have $\|({\bm Z}({\bm \Delta}))^{(k \ell)} \|_F = \| \mbox{vec}(({\bm Z}({\bm \Delta}))^{(k \ell)}) \|_2 \le 1$ for all three penalties. In the case of the lasso penalty, this property was used to prove \cite[Theorem 1]{Tugnait2024} to establish consistency in support recovery and estimation for the global minimum $\hat{\bm \Delta}$ of $L_\lambda({\bm \Delta})$. With non-convex penalties we have only a local minimum $\hat{\bm \Delta}$ satisfying (\ref{aeqn210}) with such properties. In the rest of the section we provide a proof of Theorem 1 for non-convex penalties. This requires recalling some of the developments from \cite[Appendix A]{Tugnait2024}.

In terms of $m \times m$ submatrices of ${\bm \Delta}$, $\hat{\bm \Sigma}_x$, $\hat{\bm \Sigma}_y$ and ${\bm Z}({\bm \Delta})$ corresponding to various graph edges, using $\mbox{bvec}({\bm A} {\bm D} {\bm B}) = ({\bm B}^\top \boxtimes {\bm A}) \mbox{bvec}({\bm D})$ \cite[Lemma 1]{Tracy1989}, we may rewrite (\ref{aeqn210}) as
\begin{align} 
   (\hat{\bm \Sigma}_y \boxtimes  \hat{\bm \Sigma}_x) \mbox{bvec}(\hat{\bm \Delta} )  - \mbox{bvec}(\hat{\bm \Sigma}_x-\hat{\bm \Sigma}_y) + \lambda \, \mbox{bvec}({\bm Z}(\hat{\bm \Delta}))   = {\bm 0}
  \label{aeqn220}  
\end{align}
Then (\ref{aeqn220}) can be rewritten as
\begin{align} 
  &  \begin{bmatrix} \hat{\bm \Gamma}_{S,S} & \hat{\bm \Gamma}_{S,S^c} \\
	   \hat{\bm \Gamma}_{S^c,S} & \hat{\bm \Gamma}_{S^c,S^c} \end{bmatrix} 
		 \begin{bmatrix} \mbox{bvec}(\hat{\bm \Delta}_S ) \\ \mbox{bvec}(\hat{\bm \Delta}_{S^c} ) \end{bmatrix}
	-	\begin{bmatrix} \mbox{bvec}((\hat{\bm \Sigma}_x-\hat{\bm \Sigma}_y)_S) \\ 
	  \mbox{bvec}((\hat{\bm \Sigma}_x-\hat{\bm \Sigma}_y)_{S^c} ) \end{bmatrix}  \nonumber \\
	& \quad	+ \lambda \, \begin{bmatrix} \mbox{bvec}({\bm Z}(\hat{\bm \Delta}_S)) \\
			   \mbox{bvec}({\bm Z}(\hat{\bm \Delta}_{S^c})) \end{bmatrix} = 
				\begin{bmatrix} {\bm 0} \\ {\bm 0} \end{bmatrix} \, . \label{aeqn227}  
\end{align}

The general approach of \cite{Ravikumar2011} (followed in \cite{Kolar2014, Zhang2014, Yuan2017, Jiang2018}) is to first solve the hypothetical constrained optimization problem with known edgeset $S$
\begin{equation}
  \tilde{\bm \Delta} = \arg\min_{{\bm \Delta}: {\bm \Delta}_{S^c} 
	 = {\bm 0}} L_\lambda({\bm \Delta}, \hat{\bm \Sigma}_x , \hat{\bm \Sigma}_y)  \label{aeqn250}
\end{equation}
where $S^c$ is the complement of $S$. 
Since, by construction, $\tilde{\bm \Delta}_{S^c} = {\bm 0}$, in this case (\ref{aeqn227}) reduces to 
\begin{align} 
   & \hat{\bm \Gamma}_{S,S} \mbox{bvec}(\tilde{\bm \Delta}_S )  - \mbox{bvec}((\hat{\bm \Sigma}_x-\hat{\bm \Sigma}_y)_S) + \lambda \, \mbox{bvec}({\bm Z}(\tilde{\bm \Delta}_S))    = {\bm 0} \, . \label{aeqn260}  
\end{align}
In the approach of \cite{Ravikumar2011}, one investigates conditions under which the solution $\hat{\bm \Delta}$ to (\ref{eqn20}) is the same as the solution $\tilde{\bm \Delta}$ to (\ref{aeqn250}). This is done by showing that $\hat{\bm \Delta}$ satisfies (\ref{aeqn227}). The choice $\hat{\bm \Delta} =\tilde{\bm \Delta}$ implies that $\hat{\bm \Delta}_{S^c} = {\bm 0}$ and (\ref{aeqn260}) is true with $\tilde{\bm \Delta}$ replaced with $\hat{\bm \Delta}$. In order to satisfy (\ref{aeqn227}), it remains to show that for any edge $e \in S^c$, we have strict feasibility
\begin{align} 
 \| \hat{\bm \Gamma}_{e,S} \mbox{bvec}(\tilde{\bm \Delta}_S )  - \mbox{bvec}((\hat{\bm \Sigma}_x-\hat{\bm \Sigma}_y)_e) \|_2
	       < \lambda \, , \label{aeqn261}
\end{align}
where for ${\bm a} \in \mathbb{R}^q$, $\|{\bm a} \|_2 = \sqrt{{\bm a}^\top {\bm a}}$. This requires a set of sufficient conditions stated in Theorem 1.

{\it Lemma 2} \cite{Tugnait2024}. Let $\hat{\bm \Sigma}_x$ and $\hat{\bm \Sigma}_y$ be as in (\ref{eqn10}), $\bar{\sigma}_{xy}$ as in (\ref{eqn336}), ${C}_0$ as in (\ref{eqn337}) and assume data are Gaussian. Define $n = \min(n_x, n_y)$ and
\begin{align} 
   {\mathcal A} =  & \max \big\{ \| \bm{\mathcal C}(\hat{{\bm \Sigma}_x}-{\bm \Sigma}_x^\ast) \|_\infty \, , 
	 \| \bm{\mathcal C}(\hat{{\bm \Sigma}_y}-{\bm \Sigma}_y^\ast) \|_\infty \big\}  \, . \label{aeqn310}
\end{align}
Then for any $\tau > 2$ and $n > 2m^2 \ln(4m^2 p_n^\tau)$,
\begin{align}  
   P & \Big(  {\mathcal A}  > {C}_0 \sqrt{\ln(p_n)/n} \Big) \le 2/p_n^{\tau -2} \quad \bullet \label{aeqn315}
\end{align}
Recall (\ref{eqn225})-(\ref{eqn335}) and define 
\begin{align}
  {\bm \Delta}_x = & \hat{\bm \Sigma}_x -{\bm \Sigma}_x^\ast \, , \;
	{\bm \Delta}_y = \hat{\bm \Sigma}_y -{\bm \Sigma}_y^\ast \, , \;  
	{\bm \Delta}_\Gamma =  \hat{\bm \Gamma} -{\bm \Gamma}^\ast \, , \label{aeqn400} \\
	{\bm \Delta}_\Sigma = & {\bm \Delta}_x - {\bm \Delta}_y \, , \;
	{\epsilon_0}_x =  \|\bm{\mathcal C}({\bm \Delta}_x)\|_\infty \, , \label{aeqn401} \\ 
		{\epsilon_0}_y = & \|\bm{\mathcal C}({\bm \Delta}_y)\|_\infty \, , \;
		{\epsilon_0} > \max\{ {\epsilon_0}_x, {\epsilon_0}_y \} \,.  \label{aeqn403}	
\end{align}
{\it Lemma 3} \cite{Tugnait2024}.  Assume that
\begin{align}
  \kappa_\Gamma &  < \frac{1}{3 s_n (\epsilon_0^2+2M \epsilon_0)}  \, .
    \label{aeqn405}	
\end{align}
Let (${\bm \Gamma}_{S,S}^{-\ast}$ denotes $({\bm \Gamma}_{S,S}^\ast)^{-1}$)
\begin{align}
   R({\bm \Delta}_\Gamma)= & \hat{\bm \Gamma}_{S,S}^{-1} - {\bm \Gamma}_{S,S}^{-\ast}  
	  + {\bm \Gamma}_{S,S}^{-\ast} ({\bm \Delta}_\Gamma)_{S,S} {\bm \Gamma}_{S,S}^{-\ast}\, .
    \label{aeqn406}	
\end{align}
Then we have
\begin{align}
  & \| \bm{\mathcal C}( R({\bm \Delta}_\Gamma)) \|_\infty \le  
	\frac{3}{2} \kappa_\Gamma^3 s_n ( \epsilon_0^2 + 2M \epsilon_0)^2 \, ,
    \label{aeqn407}	\\
	& \| \bm{\mathcal C}( R({\bm \Delta}_\Gamma)) \|_{1,\infty} \le  
	   \frac{3}{2} \kappa_\Gamma^3 s_n^2 ( \epsilon_0^2 + 2M \epsilon_0)^2 \, ,
    \label{aeqn408}	\\
 & \| \bm{\mathcal C}( \hat{\bm \Gamma}_{S,S}^{-1} - {\bm \Gamma}_{S,S}^{-\ast} ) \|_\infty \nonumber \\
  & \;\; \le  
	 \kappa_\Gamma^2 (\epsilon_0^2+2M \epsilon_0) \big(1+ 1.5 s_n \kappa_\Gamma (\epsilon_0^2+2M \epsilon_0) \big) \, ,
    \label{aeqn409}	\\
	&	\| \bm{\mathcal C}( \hat{\bm \Gamma}_{S,S}^{-1} - {\bm \Gamma}_{S,S}^{-\ast} ) \|_{1,\infty} \le 
	  s_n \| \bm{\mathcal C}( \hat{\bm \Gamma}_{S,S}^{-1} - {\bm \Gamma}_{S,S}^{-\ast} ) \|_\infty \quad \bullet
    \label{aeqn410}
\end{align}

Lemma 4 stated and proved below is as in \cite{Tugnait2024} but requires use of $\|({\bm Z}({\bm \Delta}))^{(k \ell)} \|_F = \| \mbox{vec}(({\bm Z}({\bm \Delta}))^{(k \ell)}) \|_2 \le 1$  where $({\bm Z}({\bm \Delta}))^{(k \ell)}$ is as in (\ref{aeqn215}). \\
{\it Lemma 4}.  Assume (\ref{aeqn405}) and the following conditions:
\begin{align}
  & 0 < \alpha < 1  \mbox{ where } \alpha \mbox{ is as in } (\ref{eqn335}) \, ,
    \label{aeqn500}	\\
	& \epsilon_0  < \min \left\{ M, \frac{ \alpha \lambda_n }{2(2-\alpha)} \right\} \, , \label{aeqn502}	\\
	& \alpha C_\alpha \min\{\lambda_n,1\}  \ge  3 s_n \epsilon_0 M \kappa_\Gamma B_s  \label{aeqn503}
\end{align}
where
\begin{align}
	C_\alpha & =  \frac{\alpha \lambda_n + 2 \epsilon_0 \alpha - 4 \epsilon_0}
	          {2 M \alpha \lambda_n + \alpha \lambda_n + 2 \epsilon_0 \alpha} \, , \label{aeqn504}	\\
	B_s & = \Big[ 1+ \kappa_\Gamma \Big( 3 s_n \epsilon_0 M + \min\{s_n M^2,M_\Sigma^2\} \Big) \nonumber \\
	  & \quad\quad \times \big( 4.5 s_n \epsilon_0 M \kappa_\Gamma +1 \big) \Big] \, . \label{aeqn505}
\end{align}
Then we have
\begin{itemize}
\item[(i)] $\mbox{bvec}(\hat{\bm \Delta}_{S^c}) = {\bm 0}$.
\item[(ii)]  $\| \bm{\mathcal C}(\hat{\bm \Delta} - {\bm \Delta}^\ast) \|_\infty 
  \le 2 \lambda_n \kappa_\Gamma + 3  s_n \epsilon_0 M \kappa_\Gamma^2 
	  \big( 4.5 s_n \epsilon_0 M \kappa_\Gamma +1 \big) \big( 2M + 2 \lambda_n \big) $ $\quad \bullet$
\end{itemize}
{\it Proof}. To establish part (i), as in \cite[Lemma 4]{Tugnait2024}, we need to show that (\ref{aeqn261}) is true. Let $d$ denote the left-side of (\ref{aeqn261}). It follow from (\ref{aeqn260}) that
\begin{align} 
  \mbox{bvec}(\tilde{\bm \Delta}_S ) = & \hat{\bm \Gamma}_{S,S}^{-1} 
	  \Big( \mbox{bvec}((\hat{\bm \Sigma}_x-\hat{\bm \Sigma}_y)_S) - \lambda \, \mbox{bvec}({\bm Z}(\tilde{\bm \Delta}_S))\Big)
		 \, . \label{aeqn510}  
\end{align}
Substitute (\ref{aeqn510}) in the left-side of (\ref{aeqn261}) to yield
\begin{align} 
  d = & \| \hat{\bm \Gamma}_{e,S} \big[ \hat{\bm \Gamma}_{S,S}^{-1} 
	  \big( \mbox{bvec}((\hat{\bm \Sigma}_x-\hat{\bm \Sigma}_y)_S) 
		 - \lambda \, \mbox{bvec}({\bm Z}(\tilde{\bm \Delta}_S))\big) \big] \nonumber \\
		& \quad - \mbox{bvec}((\hat{\bm \Sigma}_x-\hat{\bm \Sigma}_y)_e) \|_2 \, . \label{aeqn520}  
\end{align}
At the true values we have
\begin{align} 
  {\bm 0} = & \frac{\partial L({\bm \Delta}, {\bm \Sigma}_x^\ast , {\bm \Sigma}_y^\ast)}{\partial {\bm \Delta}} 
	   \Big|_{ {\bm \Delta} = {\bm \Delta}^\ast}
	     = {\bm \Sigma}_x^\ast {\bm \Delta}^\ast {\bm \Sigma}_y^\ast - ({\bm \Sigma}_x^\ast-{\bm \Sigma}_y^\ast) 
  \nonumber  
\end{align}
implying
\begin{align} 
   & {\bm \Gamma}^\ast \mbox{bvec}({\bm \Delta}^\ast )  - \mbox{bvec}({\bm \Sigma}_x^\ast-{\bm \Sigma}_y^\ast)  = {\bm 0} \, ,
  \label{aeqn530}  
\end{align}
which, noting that $({\bm \Delta}^\ast )_{S^c} = {\bm 0}$, can be rewritten as (cf.\ (\ref{aeqn227}))
\begin{align} 
  {\bm \Gamma}_{S,S}^\ast  \mbox{bvec}({\bm \Delta}_S^\ast ) = & \mbox{bvec}({\bm \Sigma}_x^\ast)_S 
	            - \mbox{bvec}({\bm \Sigma}_y^\ast)_S \, , \label{aeqn532}  \\
	{\bm \Gamma}_{e,S}^\ast  \mbox{bvec}({\bm \Delta}_S^\ast ) = & \mbox{bvec}({\bm \Sigma}_x^\ast)_e 
	            - \mbox{bvec}({\bm \Sigma}_y^\ast)_e \, . \label{aeqn534}  
\end{align}
Therefore, (${\bm A}^{-\ast} = ({\bm A}^\ast)^{-1}$),
\begin{align} 
	{\bm \Gamma}_{e,S}^\ast & {\bm \Gamma}_{S,S}^{-\ast} \big( \mbox{bvec}({\bm \Sigma}_x^\ast)_S 
	    - \mbox{bvec}({\bm \Sigma}_y^\ast)_S \big) \nonumber \\
		&	 \quad\quad -\mbox{bvec}({\bm \Sigma}_x^\ast)_e  + \mbox{bvec}({\bm \Sigma}_y^\ast)_e
							= {\bm 0}\, . \label{aeqn535}  
\end{align}
Recalling (\ref{aeqn400}) and using (\ref{aeqn535}) in (\ref{aeqn520}), 
\begin{align} 
  d = & \| \hat{\bm \Gamma}_{e,S}  \hat{\bm \Gamma}_{S,S}^{-1}  \mbox{bvec}(({\bm \Delta}_\Sigma)_S) \nonumber \\
		& + \big( \hat{\bm \Gamma}_{e,S} \hat{\bm \Gamma}_{S,S}^{-1} - {\bm \Gamma}_{e,S}^\ast {\bm \Gamma}_{S,S}^{-\ast} \big) 
		\big( \mbox{bvec}({\bm \Sigma}_x^\ast)_S - \mbox{bvec}({\bm \Sigma}_y^\ast)_S \big) \nonumber \\
		&- \lambda \, \hat{\bm \Gamma}_{e,S} \hat{\bm \Gamma}_{S,S}^{-1} 
		       \mbox{bvec}({\bm Z}(\tilde{\bm \Delta}_S))   - \mbox{bvec}(({\bm \Delta}_\Sigma)_e) \|_2 \, . \label{aeqn540}  
\end{align}
To bound various terms in (\ref{aeqn540}), using \cite[Eq.\ (80)]{Tugnait2024}, we have 
\begin{align}
& \| \hat{\bm \Gamma}_{e,S}  \hat{\bm \Gamma}_{S,S}^{-1}  \mbox{bvec}(({\bm \Delta}_\Sigma)_S) \|_2 \nonumber \\
	& \quad
 \le \|\bm{\mathcal C}( \hat{\bm \Gamma}_{e,S}  \hat{\bm \Gamma}_{S,S}^{-1} ) \|_1 \, 
      \|\bm{\mathcal C}({\bm \Delta}_\Sigma) \|_\infty \, , \label{aeqn554} \\
	& \| \big( \hat{\bm \Gamma}_{e,S} \hat{\bm \Gamma}_{S,S}^{-1} - {\bm \Gamma}_{e,S}^\ast {\bm \Gamma}_{S,S}^{-\ast} \big) 
		\big( \mbox{bvec}({\bm \Sigma}_x^\ast)_S - \mbox{bvec}({\bm \Sigma}_y^\ast)_S \big) \|_2 \nonumber \\
	& \quad  \le \|\bm{\mathcal C}( \hat{\bm \Gamma}_{e,S} \hat{\bm \Gamma}_{S,S}^{-1}
	    - {\bm \Gamma}_{e,S}^\ast {\bm \Gamma}_{S,S}^{-\ast} ) \|_1 \, 
			 \|\bm{\mathcal C}({\bm \Sigma}_x^\ast- {\bm \Sigma}_y^\ast) \|_\infty \, , \label{aeqn555} \\
& \|  \hat{\bm \Gamma}_{e,S} \hat{\bm \Gamma}_{S,S}^{-1}  \mbox{bvec}({\bm Z}(\tilde{\bm \Delta}_S)) \|_2 \nonumber \\
	& \quad
 \le \|\bm{\mathcal C}( \hat{\bm \Gamma}_{e,S} \hat{\bm \Gamma}_{S,S}^{-1} ) \|_1 \, 
      \|\bm{\mathcal C}({\bm Z}(\tilde{\bm \Delta}_S)) \|_\infty  \nonumber \\
& \quad \le  \|\bm{\mathcal C}( \hat{\bm \Gamma}_{e,S} \hat{\bm \Gamma}_{S,S}^{-1} ) \|_1 \, , \label{aeqn556}	\\
& \| \mbox{bvec}(({\bm \Delta}_\Sigma)_e) \|_2 
 \le \|\bm{\mathcal C}({\bm \Delta}_\Sigma) \|_\infty \, , \label{aeqn557}	
\end{align}
where in (\ref{aeqn556}) we have used the fact that
\[
  \|\bm{\mathcal C}({\bm Z}(\tilde{\bm \Delta}_S)) \|_\infty 
	    = \max_{\{k, \ell\} \in S} \| ({\bm Z}(\tilde{\bm \Delta}_S))^{(k, \ell)} \|_F \le 1 \, .
\]
Now mimic the proof of \cite[Lemma 4]{Tugnait2024} from \cite[Eq.\ (85)]{Tugnait2024} through \cite[Eq.\ (101)]{Tugnait2024} to conclude that $d < \lambda$, and hence, (\ref{aeqn261}) is true, proving part (i) of Lemma 4.

We now turn to the proof of Lemma 4(ii). Since $\hat{\bm \Delta} = \tilde{\bm \Delta}$, for any edge $\{k, \ell \} \in S$, we have 
\begin{align}
	& \| (\hat{\bm \Delta} - {\bm \Delta}^\ast )^{(k \ell)} \|_F 
	   = \| (\tilde{\bm \Delta} - {\bm \Delta}^\ast )^{(k \ell)} \|_F \nonumber \\
  & \quad
	  = \| \mbox{vec}(\tilde{\bm \Delta}^{(k \ell)}) - \mbox{vec}(({\bm \Delta}^{\ast})^{(k \ell)}) \|_2
		 \, .  \label{aeqn590}	
\end{align}
Using (\ref{aeqn260}) and (\ref{aeqn532})
\begin{align}
	& \mbox{bvec}( (\tilde{\bm \Delta} - {\bm \Delta}^\ast )_S ) 
	= \hat{\bm \Gamma}_{S,S}^{-1} \mbox{bvec}(({\bm \Delta}_\Sigma)_S) 
	  + ( \hat{\bm \Gamma}_{S,S}^{-1} - {\bm \Gamma}_{S,S}^{-\ast} )\nonumber \\
  & \quad \times 
	 \mbox{bvec}(({\bm \Sigma}_x^\ast - {\bm \Sigma}_y^\ast)_S) 
	  - \lambda_n \hat{\bm \Gamma}_{S,S}^{-1} \, \mbox{bvec}({\bm Z}(\tilde{\bm \Delta}_S))
		 \, .  \label{aeqn592}	
\end{align}
Since $ \hat{\bm \Gamma}_{S,S}^{-1} = \hat{\bm \Gamma}_{S,S}^{-1} - {\bm \Gamma}_{S,S}^{-\ast} +{\bm \Gamma}_{S,S}^{-\ast} \, ,$ 
\begin{align}
	& \|\bm{\mathcal C}( \hat{\bm \Gamma}_{S,S}^{-1} ) \|_{1,\infty} \le 
	  \|\bm{\mathcal C}( \hat{\bm \Gamma}_{S,S}^{-1} - {\bm \Gamma}_{S,S}^{-\ast}  ) \|_{1,\infty} +
	\|\bm{\mathcal C}( {\bm \Gamma}_{S,S}^{-\ast} ) \|_{1,\infty}
		 \, .  \label{aeqn594}	
\end{align}
By (\ref{aeqn592}), for any edge $f = \{k, \ell\} \in S$, we have
\begin{align}
	& \| \mbox{vec}( (\tilde{\bm \Delta} - {\bm \Delta}^\ast)^{(k \ell)}  ) \|_2
	\le \| (\hat{\bm \Gamma}_{S,S}^{-1} - {\bm \Gamma}_{S,S}^{-\ast})_{f,S}  \nonumber \\
  & \quad\quad\quad \times
	  \mbox{bvec} \big( ({\bm \Delta}_\Sigma)_S + ({\bm \Sigma}_x^\ast - {\bm \Sigma}_y^\ast)_S
		- \lambda_n {\bm Z}(\tilde{\bm \Delta}_S) \big) \|_2  \nonumber \\
  & \quad + \| ({\bm \Gamma}_{S,S}^{-\ast})_{f,S} \,
	 \mbox{bvec} \Big( ({\bm \Delta}_\Sigma)_S  
		- \lambda_n {\bm Z}(\tilde{\bm \Delta}_S) \Big) \|_2  \nonumber \\
	&  \le \|(\bm{\mathcal C}( \hat{\bm \Gamma}_{S,S}^{-1} - {\bm \Gamma}_{S,S}^{-\ast}  ))_{f,S} \|_{1} \,
	  \Big( \|\bm{\mathcal C}( {\bm \Delta}_\Sigma ) \|_{\infty} 
		 + \|\bm{\mathcal C}( {\bm \Sigma}_x^\ast - {\bm \Sigma}_y^\ast ) \|_{\infty}   \nonumber \\
  & \quad + \lambda_n \|\bm{\mathcal C}({\bm Z}(\tilde{\bm \Delta}_S)) \|_\infty \Big) 
	   + \|(\bm{\mathcal C}( {\bm \Gamma}_{S,S}^{-\ast} ) )_{f,S}\|_{1} \,
	   \big( \|\bm{\mathcal C}( {\bm \Delta}_\Sigma ) \|_{\infty} \nonumber \\
  & \quad
		   + \lambda_n \|\bm{\mathcal C}({\bm Z}(\tilde{\bm \Delta}_S)) \|_\infty \big)  \nonumber \\
  &  \le \|\bm{\mathcal C}( \hat{\bm \Gamma}_{S,S}^{-1} - {\bm \Gamma}_{S,S}^{-\ast}  ) \|_{1,\infty} \,
	  \Big( \|\bm{\mathcal C}( {\bm \Delta}_\Sigma ) \|_{\infty} 
		 + \|\bm{\mathcal C}( {\bm \Sigma}_x^\ast - {\bm \Sigma}_y^\ast ) \|_{\infty}   \nonumber \\
  & \quad\quad + \lambda_n \Big) + \|\bm{\mathcal C}( {\bm \Gamma}_{S,S}^{-\ast} ) \|_{1,\infty} \,
	   \big( \|\bm{\mathcal C}( {\bm \Delta}_\Sigma ) \|_{\infty} + \lambda_n \big)  \nonumber \\
  &  \le s_n \kappa_\Gamma^2 (\epsilon_0^2 + 2M \epsilon_0)  
	      \big(1+1.5 s_n (\epsilon_0^2 + 2M \epsilon_0) \kappa_\Gamma \big)  \nonumber \\
  &  \quad\quad\quad \times (2 \epsilon_0 + 2 M + \lambda_n) + \kappa_\Gamma (2 \epsilon_0 + \lambda_n)  =: U_{b3}
		 \, .  \label{aeqn596}	
\end{align}
By (\ref{aeqn502}), for $0 < \alpha < 1$, we have $2 \epsilon_0 < \alpha \lambda_n /(2-\alpha) < \alpha \lambda_n < \lambda_n$. Therefore, $\kappa_\Gamma (2 \epsilon_0 + \lambda_n) < 2 \kappa_\Gamma \lambda_n$ and $2 \epsilon_0 + 2 M + \lambda_n < 2M + 2 \lambda_n$. Since $\epsilon_0 < M$ by (\ref{aeqn502}), we also have $\epsilon_0^2 + 2M \epsilon_0 < 3M \epsilon_0$. Using these relations and (\ref{aeqn596}), it follows that
\begin{align*}
	U_{b3} \; \le \; &  3  s_n \epsilon_0 M \kappa_\Gamma^2 
	  \big( 1+ 4.5 s_n \epsilon_0 M \kappa_\Gamma  \big) \big( 2M + 2 \lambda_n \big) + 2 \lambda_n \kappa_\Gamma 
		 \, .  
\end{align*}
Finally, 
\begin{align*}
  \| \bm{\mathcal C}(\hat{\bm \Delta} - {\bm \Delta}^\ast) \|_\infty = & \max_{f = \{k, \ell\} \in S} \| \mbox{vec}( (\tilde{\bm \Delta} - {    \bm \Delta}^\ast)^{(k \ell)}  ) \|_2 \, ,
\end{align*}
proving the desired result. $\quad \blacksquare$

{\it Proof of Theorem 1}. With Lemmas 2-4 in place, we simply mimic the proof of \cite[Theorem 1]{Tugnait2024}; no changes are needed. $\quad \blacksquare$

\section{Proofs of Lemma 1 and Theorem 2} \label{append2}

{\it Proof of Lemma 1}. We first prove part (i). Following Eqns.\ (135)-(137) of \cite{Tugnait2024} we have
\begin{align}
 & |\tilde{\bm \theta}^\top  (\hat{\bm \Gamma}-{\bm \Gamma}^\ast) \tilde{\bm \theta}| \le
    16 s_n (\epsilon_0^2 + 2 M \epsilon_0) \| \tilde{\bm \theta} \|_2^2 \, . \label{aeqn750}
\end{align}
We have
\begin{align}
 \tilde{\bm \theta}^\top \hat{\bm \Gamma} \tilde{\bm \theta} = & \tilde{\bm \theta}^\top  {\bm \Gamma}^\ast \tilde{\bm \theta}
	   + \tilde{\bm \theta}^\top  (\hat{\bm \Gamma}-{\bm \Gamma}^\ast) \tilde{\bm \theta} \nonumber \\
  \ge & \phi_{min}^\ast \, \| \tilde{\bm \theta} \|_2^2 -
	   48 s_n  M \epsilon_0 \,  \| \tilde{\bm \theta} \|_2^2  \label{aeqn752}
\end{align}
where we used the fact that $\epsilon_0 < M$, by Lemma 4. Pick $\epsilon_0$ to satisfy
\begin{align}
 \epsilon_0 = & C_0 \sqrt{ln(p_n)/n} \; \le \; \min \big\{ M, \frac{\phi_{min}^\ast}{192 s_n M} \big\}
  \, ,  \label{aeqn754}
\end{align}
leading to $48 s_n  M \epsilon_0 \le \phi_{min}^\ast/4$ w.h.p.\ for $n > N_2$. This proves Lemma 1(i). We now turn to part (ii). 
With ${\bm \Delta}_\Gamma =  \hat{\bm \Gamma} -{\bm \Gamma}^\ast$ and ${\bm \theta}$ such that $\mbox{supp}({\bm \theta}) \subseteq \mbox{supp}({\bm \theta}^\ast)$, we have
\begin{align}
{\bm \theta}_S^\top \hat{\bm \Gamma}_{S,S} {\bm \theta}_S = &
	  {\bm \theta}^\top \hat{\bm \Gamma} {\bm \theta}
		 = {\bm \theta}^\top {\bm \Gamma}^\ast {\bm \theta} + 
		  {\bm \theta}^\top {\bm \Delta}_\Gamma {\bm \theta} \, . \label{aeqn756}
\end{align}
As in Eqn.\ (135) of \cite{Tugnait2024}, we have
\begin{align}
  {\bm \theta}^\top {\bm \Delta}_\Gamma {\bm \theta} = &
    \sum_{t_1=1}^{p^2} \sum_{t_2=1}^{p^2} {\bm \theta}^\top_{Gt_1} ({\bm \Delta}_\Gamma)_{Gt_1,Gt_2} 
		       {\bm \theta}_{Gt_2}
  \, . \label{aeqn758}
\end{align}
Therefore,
\begin{align}
 & |{\bm \theta}^\top {\bm \Delta}_\Gamma {\bm \theta}| \le
    \sum_{t_1=1}^{p^2} \sum_{t_2=1}^{p^2} |{\bm \theta}^\top_{Gt_1} ({\bm \Delta}_\Gamma)_{Gt_1,Gt_2} {\bm \theta}_{Gt_2}| \nonumber \\
	& \quad  \le \sum_{t_1=1}^{p^2} \sum_{t_2=1}^{p^2} \|{\bm \theta}_{Gt_1}\|_2 \,
	  \|({\bm \Delta}_\Gamma)_{Gt_1,Gt_2} \|_F \, \|{\bm \theta}_{Gt_2}\|_2  \nonumber \\
	& \quad  \le \| \bm{\mathcal C}({\bm \Delta}_\Gamma )\|_\infty 
	 \big( \sum_{t=1}^{p^2}  \|{\bm \theta}_{Gt}\|_2  \big)^2  \nonumber \\
	& \quad  \le (\epsilon_0^2 + 2 M \epsilon_0) \, s_n \|{\bm \theta}\|_2^2   \le 3 M \epsilon_0 s_n \|{\bm \theta}\|_2^2
      \, ,  \label{aeqn760}
\end{align}
where we used $\| \bm{\mathcal C}({\bm \Delta}_\Gamma )\|_\infty < \epsilon_0^2 + 2 M \epsilon_0$ by \cite[Eqn.\ (57)]{Tugnait2024},  $\epsilon_0 < M$ (Lemma 4) and the fact that by the Cauchy-Schwarz inequality, 
\[
   \sum_{t=1}^{p^2}  \|\tilde{\bm \theta}_{Gt}\|_2 \le \sqrt{s_n} \, \|{\bm \theta}\|_2 \, .
\]
Thus, for ${\bm \theta}$ such that $\mbox{supp}({\bm \theta}) \subseteq \mbox{supp}({\bm \theta}^\ast)$, 
\begin{align}
{\bm \theta}_S^\top \hat{\bm \Gamma}_{S,S} {\bm \theta}_S \ge &
	   \phi_{min}^\ast \, \| {\bm \theta} \|_2^2 -
	   3 s_n  M \epsilon_0 \,  \| {\bm \theta} \|_2^2 
		\ge \frac{63}{64} \phi^\ast_{\min} \|{\bm \theta}_S\|_2^2 \label{aeqn764}
\end{align}
for $n > N_2$ with $\epsilon_0$ chosen as in part (i) of Lemma 1. $\quad \blacksquare$

{\it Proof of Theorem 2}. With ${\bm \Delta}$ restricted to $\mbox{supp}({\bm \Delta}) \subseteq \mbox{supp}({\bm \Delta}^\ast)$, by Lemma 1(ii), $L({\bm \Delta}) - \frac{\mu}{2} \| {\bm \Delta} \|_F^2 = L({\bm \Delta}_S) - \frac{\mu}{2} \| {\bm \Delta}_S \|_F^2$ is strictly convex for $\mu < (63/64) \phi^\ast_{\min}$ since its Hessian is $\hat{\bm \Gamma}_{S,S} - \mu {\bm I}_{m^2 s_n}$. By property (v) of the penalty functions, $g(u):=\rho_\lambda(u) +\frac{\mu}{2} u^2$ is convex, for some $\mu \ge 0$ ($\mu=0$ for lasso, $=1/(1-a)$ for SCAD, and $=\lambda_n/\epsilon$ for log-sum), and by property (ii), it is non-decreasing on $\mathbb{R}_+$. Therefore, by the composition rules \cite[Sec.\ 3.2.4]{Boyd2004}, $g(\| {\bm \Delta}_S^{(k \ell )} \|_F)$ is convex. Hence, under (\ref{eqn810}), $L_\lambda({\bm \Delta}_S)$ is strictly convex in ${\bm \Delta}_S$ and therefore, the solution to (\ref{aeqn250}) via (\ref{aeqn210})-(\ref{aeqn227}) and (\ref{aeqn260}) yields a unique minimizer. Finally, by Theorem 1(ii), unrestricted $\hat{\bm \Delta}$ of Theorem 1 is unique.
 $\quad \blacksquare$

\bibliographystyle{unsrt} 

\begin{thebibliography}{}

\end{thebibliography}


\begin{thebibliography}{00}

{

\bibitem{Whittaker1990} J.\ Whittaker, {\it Graphical Models in Applied Multivariate Statistics}. New York: Wiley, 1990.

\bibitem{Lauritzen1996} S.L.\ Lauritzen, {\it Graphical models}. Oxford, UK: Oxford Univ.\ Press, 1996.

\bibitem{Buhlmann2011} P.\ B\"{u}hlmann and S.\ van de Geer, {\em Statistics for High-Dimensional data}. Berlin: Springer, 2011.

\bibitem{Lam2009} C.\ Lam and J.\ Fan, ``Sparsistency and rates of convergence in large covariance matrix estimation,'' {\em Ann.\ Statist.}, vol.\ 37, no.\ 6B, pp.\ 4254-4278, 2009.

\bibitem{Ravikumar2011} P.\ Ravikumar, M.J.\ Wainwright, G.\ Raskutti and B.\ Yu, ``High-dimensional covariance estimation by minimizing $\ell_1$-penalized log-determinant divergence,'' {\em Electronic J.\ Statistics}, vol.\ 5, pp.\ 935-980, 2011.

\bibitem{Wainwright2019} M.J.\ Wainwright, {\em High-Dimensional Statistics: A Non-Asymptotic Viewpoint}. Cambridge, UK: Cambridge Univ.\ Press, 2019.

\bibitem{Yuan2017} H.\ Yuan, R.\ Xi, C.\ Chen and M.\ Deng, ``Differential network analysis via lasso penalized D-trace loss,'' {\it Biometrika}, vol.\ 104, pp.\ 755-770, 2017.

\bibitem{Tang2020} Z.\ Tang, Z.\ Yu and C.\ Wang, ``A fast iteraive algorithm for high-dimensional differential network,'' {\it Computational Statistics}, vol.\ 35, pp.\ 95-109, 2020.

\bibitem{Wu2020} Y.\ Wu, T.\ Li, X.\ Liu and L.| Chen, ``Differential network inference via the fused D-trace loss with cross variables,'' {\it Electronic J.\ Statistics}, vol.\ 14, pp.\ 1269-1301, 2020.

\bibitem{Zhao2014} S.D.\ Zhao, T.T.\ Cai and H.\ Li, ``Direct estimation of differential networks,'' {\it Biometrika}, vol.\ 101, pp.\ 253-268, June 2014.

\bibitem{Belilovsky2016} E.\ Belilovsky,  G.\ Varoquaux and M.B.\ Blaschko, ``Hypothesis testing for differences in Gaussian graphical models: Applications to brain connectivity,'' {\it Advances in Neural Information Processing Systems (NIPS 2016)}, vol.\ 29, Dec.\ 2016.

\bibitem{Danaher2014} P.\ Danaher, P.\ Wang and D.M.\ Witten, ``The joint graphical lasso for
inverse covariance estimation across multiple classes,'' {\it J.\ Royal Statistical
Society, Series B (Methodological)}, vol.\ 76, pp.\ 373-397, 2014.

\bibitem{Kolar2014} M.\ Kolar, H.\ Liu and E.P.\ Xing, ``Graph estimation from multi-attribute data,'' {\em J.\ Machine Learning Research}, vol.\ 15, pp.\ 1713-1750, 2014.

\bibitem{Tugnait21a} J.K.\ Tugnait, ``Sparse-group lasso for graph learning from multi-attribute data,'' {\it IEEE Trans.\ Signal Process.}, vol.\ 69, pp.\ 1771-1786, 2021. (Corrections, vol.\ 69, p.\ 4758, 2021.)

\bibitem{Tugnait2024} J.K.\ Tugnait, ``Learning high-dimensional differential graphs from multi-attribute data,'' {\em IEEE Trans.\ Signal Process.}, vol.\ 72, pp.\ 415-431, 2024. 

\bibitem{Marjanovic18} G.\ Marjanovic and V.\ Solo, ``Vector $l_0$ sparse conditional independence graphs,'' in {\it Proc.\ IEEE ICASSP 2018}, pp.\ 2731-2735, 2018.

\bibitem{Sundaram20}  P.\ Sundaram, M.\ Luessi, M.\ Bianciardi, S.\ Stufflebeam, M. H\"am\"al\"ainen and V.\ Solo, ``Individual resting-state brain networks enabled by massive multivariate conditional mutual information,'' {\it IEEE Trans.\ Med.\ Imaging}, vol.\ 39, pp. 1957-1966, 2020.

\bibitem{Zhao2022} B.\ Zhao, Y.S.\ Wang and M.\ Kolar, ``FuDGE: A method to estimate a functional differential graph in a high-dimensional setting,'' {\it J.\ Machine Learning Research}, vol.\ 23, pp.\ 1-82, 2022.

\bibitem{Fan2001} J.\ Fan and R.\ Li, ``Variable selection via nonconcave penalized likelihood and its
oracle properties,'' {\it J.\ American Statistical Assoc.}, vol.\ 96, pp.\ 1348-1360, Dec.\ 2001. 

\bibitem{Candes2008} E.J.\ Cand\`{e}s, M.B.\ Wakin and S.P.\ Boyd, ``Enhancing sparsity by reweighted $\ell_1$ minimization,'' {\it J.\ Fourier Anal.\ Appl.}, vol.\ 14, pp.\  877-905, 2008.
	
\bibitem{Zou2008} H.\ Zou and R.\ Li, ``One-step sparse estimates in nonconcave penalized likelihood models,'' {\em Ann.\ Statist.}, vol.\ 36, no.\ 4, pp.\ 1509-1533, 2008.

\bibitem{Tugnait21b} J.K.\ Tugnait, ``Sparse graph learning under Laplacian-related constraints,'' {\em IEEE Access}, vol.\ 9, pp.\ 151067-151079, 2021.

\bibitem{Tugnait22} J.K.\ Tugnait, ``Sparse-group log-sum penalized graphical model learning for time series,'' in {\it Proc.\  2022 IEEE Intern.\ Conf.\ Acoustics, Speech \& Signal Processing (ICASSP 2022)}, pp.\ 5822-5826, Singapore, May 22-27, 2022.

\bibitem{Wei2023} Q.\ Wei and Z.\ Zhao, ``Large covariance matrix estimation with oracle statistical rate via majorization-minimization," {\it IEEE Trans.\  Signal Proc.}, vol.\ 71, pp.\ 3328-3342, 2023.

\bibitem{Varma2020} R.\ Varma, H.\ Lee, J.\ Kova\v{c}evi\'{c} and Y.\ Chi, ``Vector-valued graph trend filtering with non-convex penalties,'' {\it IEEE Trans.\ Signal and Inform.\ Proc.\ over Networks}, vol.\ 6, pp.\ 48-62, 2020.

\bibitem{Xu16} P.\ Xu and Q.\ Gu, ``Semiparametric differential graph models,'' in {\it Proc.\ 30th Conference on Neural Information Processing Systems (NIPS 2016)}, Barcelona, Spain, 2016.

\bibitem{Tugnait23} J.K.\ Tugnait, ``Estimation of differential graphs via log-sum penalized D-trace loss,'' in {\it Proc.\ 22nd IEEE Statistical Signal Processing Workshop (SSP-2023)}, pp.\ 240-244, Hanoi, Vietnam, July 2-5, 2023. 

\bibitem{Jiang2018} B.\ Jiang, X.\ Wang and C.\ Leng, ``A direct approach for sparse quadratic discriminant analysis,'' {\it J.\ Machine Learning Research}, vol.\ 19, pp.\ 1-37, 2018.

\bibitem{Zhang2014} T.\ Zhang and H.\ Zou, ``Sparse precision matrix estimation via lasso penalized D-trace loss,'' {\it Biometrika}, vol.\ 101, pp.\ 103-120, 2014.

\bibitem{Kumar2019} S.\ Kumar, J.\ Ying, J.V.\ de Miranda Cardoso, and D.\ Palomar, ``Structured graph learning via Laplacian spectral constraints,'' in {\it Proc.\ 32nd Conference on Neural Information Processing Systems (NeurIPS 2019)}, Vancouver, Canada, 2019.

\bibitem{Medvedovsky2024} Y.\ Medvedovsky, E.\ Treister and  T.S.\ Routtenberg, ``Efficient graph Laplacian estimation by proximal Newton,'' in {\it Proc.\ 27th Intern.\ Conf.\ Artificial Intelligence and Statistics (AISTATS 2024), PMLR}, vol.\  238, pp.\ 1171-1179, Valencia, Spain, May 2024.

\bibitem{Tracy1989} D.S.\ Tracy and K.G.\ Jinadasa, ``Partitioned Kronecker products of matrices and applications,'' {\it Canadian J.\ Statistics}, vol.\ 17, pp.\ 107-120, March 1989.

\bibitem{Liu2008} S.\ Liu, ``Matrix results on Khatri-Rao and Tracy-Singh products,'' {\it Linear Algebra \& Its Applications}, vol.\ 289, pp.\ 267-277, 1999.

\bibitem{Yuan2006} M.\ Yuan and Y.\ Lin, ``Model selection and estimation in regression with grouped variables,'' {\em J.\ Royal Statistical Society, Series B (Methodological)}, vol.\ 68, pp.\ 49-67, 2006.

\bibitem{Loh2017}  P.-L.\ Loh and M.J.\ Wainwright, ``Support recovery without incoherence: A case for non-convex
regularization,'' {\it Ann.\ Statist.}, vol.\ 45, pp.\ 2455-2482, 2017.

\bibitem{Beck2009} A.\ Beck and M.\ Teboulle, ``A fast iterative shrinkage-thresholding algorithm for linear inverse problems,'' {\it SIAM Journal on Imaging Sciences}, vol.\ 2, no.\ 1,  pp.\ 183-202, 2009.

\bibitem{Boyd2014} N.\ Parikh and N.P.\ Boyd, ``Proximal algorithms,'' {\it Foundations and Trends in Machine Learning}, vol.\ 1, no.\ 3, pp.\ 127-239, 2014.

\bibitem{Yao2018} Q.\ Yao and J.T.\ Kwok, ``Efficient learning with a family of non-convex regularizers by
redistributing non-convexity,'' {\it J.\ Machine Learning Research}, vol.\ 18, pp.\ 1-52, 2018.

\bibitem{Gong2013} P.\ Gong, C.\ Zhang, Z.\ Lu, J.\ Huang and J.\ Ye, ``A general iterative shrinkage and thresholding algorithm for non-convex regularized optimization problems,'' in {\it Proc.\ 30th Intern.\ Conf.\ Machine Learning (ICML)}, pp.\ 37–45, 2013.

\bibitem{Negahban2012} S.N.\ Negahban, P.\ Ravikumar, M.J.\ Wainwright and B.\ Yu, ``A unified framework for high-dimensional analysis of M-estimators with decomposable regularizers,'' {\it Statistical Science}, vol.\ 27, No.\ 4, pp.\ 538-557, 2012.

\bibitem{Loh2015} P.-L.\ Loh and M.J.\ Wainwright, ``Regularized M-estimators with non-convexity: Statistical and
algorithmic theory for local optima,'' {\it J.\ Machine Learning Research}, vol.\ 16, pp.\ 559-616, 2015.
	
\bibitem{Bertsekas} D.P.\ Bertsekas, A.\ Nedi\'{c} and A.E.\ Ozdaglar, {\em Convex Analysis and Optimization}, Belmont, MA: 
   Athena Scientific, 2003.
	
\bibitem{Barabasi1999} A-L.\ Barab\'{a}si and R.\ Albert, R\'{e}ka, ``Emergence of scaling in random networks,'' {\it Science}, vol.\ 286, no.\ 5439, pp.\ 509-512, Oct.\ 1999.

\bibitem{Lu2014} S.\ Lu, J.\ Kang, W.\ Gong and D.\ Towsley, ``Complex network comparison using random walks,'' in {\it WWW '14 Companion: Proc.\ 23rd Intern.\ Conf.\ World Wide Web}, pp.\ 727-730, Seoul, Korea, April 2014.

\bibitem{Zhang2017} S.\ Zhang, B.\ Guo, A.\ Dong, J.\ He, Z.\  Xu and S.X.\ Chen, ``Cautionary tales on air-quality improvement in Beijing,'' {\em Proc.\ Royal Society A: Mathematical, Physical and Engineering Sciences}, vol.\ 473, p.\ 20170457, 2017.

\bibitem{Chen2015} W.\ Chen, F.\ Wang, G.\ Xiao, J.\ Wu and S.\ Zhang, ``Air quality of Beijing and impacts of the new ambient air
quality standard,'' {\em Atmosphere}, vol.\ 6, pp.\ 1243-1258, 2015.

\bibitem{Boyd2004} S.\ Boyd and L.\ Vandenberghe, {\em Convex Optimization}, Cambridge, UK: 
   Cambridge Univ.\ Press, 2004.

}
          
\end{thebibliography}

\end{document}